\documentclass[preprint,12pt]{elsarticle}

\usepackage{hyperref}       
\usepackage{url}            
\usepackage{booktabs}       
\usepackage{amsfonts}       
\usepackage{nicefrac}       
\usepackage{microtype}      
\usepackage{adjustbox}
\usepackage{subcaption}

\usepackage{amssymb,amsmath,amsthm,amsfonts}
\usepackage{framed}
\usepackage{mdframed}
\usepackage[dvipsnames]{xcolor}
\usepackage{enumitem}
\usepackage{tikz}
\usetikzlibrary{calc}
\usepackage{relsize} 

\usepackage{todonotes}

\newcommand*{\tran}{^{\mkern-0.5mu\mathsf{T}}}
\newcommand*{\inv}{^{\mkern-0.5mu\mathsf{-1}}}

\newcommand*{\Norm}{\mathcal{N}}

\newcommand{\solveopt}[2]{\underset{#2}{\textup{#1}}}
\newcommand{\argmax}[1]{\solveopt{argmax}{#1}}

\newcommand{\eqdef}{\overset{\mathsmaller{\triangle}}{=}}

\newcommand{\ROCAUC}{ROC AUC}

\newcommand{\concMF}{C}
\newcommand{\stackedMF}{S}
\newcommand{\xgb}{xgb}
\newcommand{\logit}{logit}
\newcommand{\gpc}{gpc}
\newcommand{\mfgpc}{MF gpc}
\newcommand{\gpma}{gp-ma}
\newcommand{\hetmogp}{hetmogp}

\newcommand{\diabetes}{dbts}
\newcommand{\german}{grmn}
\newcommand{\waveform}{wvfr}
\newcommand{\satimage}{stmg}
\newcommand{\splice}{splc}
\newcommand{\spambase}{spmb}
\newcommand{\hypothyroid}{hpth}
\newcommand{\mushroom}{mshr}
\newcommand{\musicgenre}{mscg}
\newcommand{\sentimentpolarity}{sntp}

\definecolor{mfgpc}{RGB}{0, 0, 0}
\definecolor{hetmogp}{RGB}{255, 165, 0}
\definecolor{gpc}{RGB}{228, 26, 28}
\definecolor{logit}{RGB}{77, 175, 74}
\definecolor{major_vote}{RGB}{149, 149, 149}
\definecolor{stacked_mf_gpc_x3}{RGB}{228, 26, 28}
\definecolor{stacked_mf_logit_x3}{RGB}{77, 175, 74}
\definecolor{stacked_mf_xgb_x3}{RGB}{55, 126, 184}
\definecolor{conc_mf_gpc_x3}{RGB}{228, 26, 28}
\definecolor{conc_mf_logit_x3}{RGB}{77, 175, 74}
\definecolor{conc_mf_xgb_x3}{RGB}{55, 126, 184}
\definecolor{xgb}{RGB}{55, 126, 184}

\definecolor{budget_noise_0.0}{RGB}{31, 119, 180}
\definecolor{budget_noise_0.2}{RGB}{255, 127, 14}
\definecolor{budget_noise_0.3}{RGB}{44, 160, 44}
\definecolor{budget_noise_0.4}{RGB}{214, 39, 40}

\definecolor{budget_mfgpc}{RGB}{64, 64, 64}
\definecolor{budget_gpc}{RGB}{187, 20, 25}
\definecolor{gpma}{RGB}{255, 0, 255}

\newcommand{\dashdottedline}[1]{\begin{tikzpicture}[baseline=-0.5ex]
                                \draw [thick,dash dot, #1] (0,0) -- (0.75,0);
                                \end{tikzpicture}}
                                
\newcommand{\dashedline}[1]{\begin{tikzpicture}[baseline=-0.5ex]
                                \draw [thick,dashed, #1] (0,0) -- (0.75,0);
                                \end{tikzpicture}}
                                
\newcommand{\dottedline}[1]{\begin{tikzpicture}[baseline=-0.5ex]
                                \draw [thick,dotted, #1] (0,0) -- (0.75,0);
                                \end{tikzpicture}}
                                
\newcommand{\plainline}[1]{\begin{tikzpicture}[baseline=-0.5ex]
                                \draw [thick, #1] (0,0) -- (0.75,0);
                                \end{tikzpicture}}

\journal{Neurocomputing}

\begin{document}

\begin{frontmatter}



\title{Gaussian Process Classification for Variable Fidelity Data}

\author[SKT_address,correspondingauthor]{Nikita Klyuchnikov}
\cortext[correspondingauthor]{Corresponding author}
\ead{nikita.klyuchnikov@skolkovotech.ru}
\author[SKT_address]{Evgeny Burnaev}
\address[SKT_address]{Skolkovo Institute of Science and Technology, Skolkovo Innovation Center, Building 3,
Moscow  143026, Russia}

\begin{abstract}
In this paper we address a classification problem where two sources of labels with different levels of fidelity are available. Our approach is to combine data from both sources by applying a co-kriging schema on latent functions, which allows the model to account item-dependent labeling discrepancy. We provide an extension of Laplace inference for Gaussian process classification, that takes into account multi-fidelity data. We evaluate the proposed method on real and synthetic datasets and show that it is more resistant to different levels of discrepancy between sources than other approaches for data fusion. Our method can provide accuracy/cost trade-off for a number of practical tasks such as crowd-sourced data annotation and feasibility regions construction in engineering design.
\end{abstract}


\end{frontmatter}


\section{Introduction}
\label{sec:introduction}

The problem of multi-fidelity modeling \citep{peherstorfer2016survey} arises in the broad range of applied disciplines, such as engineering design, medical diagnostics, and even product development, when an object of interest can be modeled with a cheaper, yet typically less reliable alternative. The main motivation behind multi-fidelity modeling is that low-fidelity data can bring additional benefits in terms of accuracy/cost trade-off, when it is used properly along with high-fidelity data \citep{pmlr-v54-zaytsev17a,Toal:2015:CRU:2790713.2790736}. For example, an article \cite{snow2008cheap} demonstrates that high-quality linguistic annotation results can be achieved with much lower expenses when non-expert annotators (i.e. low-fidelity data) are employed. The authors concluded that four non-experts per item were enough on average to achieve an expert-level annotation quality for their tasks, although this condition can be relaxed further, by requiring multiple annotations only for a fraction of the dataset. Similarly, in engineering design \citep{PAWLUS20115091} a high-fidelity source of data can be a physical experiment, whereas a low-fidelity can be a mathematical model or a computer simulation. 

Multi-fidelity modeling based on Gaussian processes (GPs) \citep{Rasmussen2006gaussianprocesses} is a reasonable approach for applications discussed above, because of the Bayesian formulation, which allows incorporation of the prior knowledge about the task into the prediction and makes learning on small samples more robust. The latter is especially important, since high-fidelity data typically contains just a few examples. In addition, Gaussian processes are based on kernel functions, whose hyperparameters can be selected via marginal likelihood maximization instead of grid search with cross-validation. 

Gaussian process regression for multi-fidelity data has been thoroughly studied in recent years \citep{forrester2007multi,Zaytsev2017}, however multi-fidelity classification based on Gaussian processes has been left behind until recently. For example, the work about feasibility regions for aeroelastic stability modeling \cite{dribusch2013multi} pointed out that multi-fidelity methods had been limited to continuous response models. Although discrete response models can also be approximated with continuous ones, in some extreme cases, such as binary classification, continuous approximations seem as weird as using Linear regression instead of Logistic regression. On the other hand, developing appropriate models for multi-fidelity classification is essential, because there are problems in engineering design with discrete responses. 
For instance, report \cite{mouret201720} points out the problem of reality gap in robotic simulators and argues the importance of their ability to estimate reliability regions, where accomplishment of actions is accurately predicted by the simulator. This problem has binary responses i.e. success or fail; simulated outcomes of robot's actions are low-fidelity data, whereas observations of real executions are high-fidelity data. Furthermore, discrete responses are common and convenient when the object of interest is a human. For example, users say they either like a new feature of the product or not during A/B testing, which gives direct evidence of their attitude i.e. high-fidelity data, or users are just asked to imagine the feature and express their preferences during interviewing i.e. low-fidelity data. 

In this work, we propose a co-kriging model for latent low- and high- fidelity functions and extend the Laplace inference algorithm for Gaussian process classification to handle this case. The novelty of our work with respect to other existing ones is adaptation of co-kriging model to classification problem. This model imposes specific dependency and order on sources of data, which help it achieve better performance that more general methods in cases when nature of data is well explained by the model. We evaluate the proposed method on three groups of datasets: artificially generated under the model assumptions, real benchmark datasets with simulated noise for low-fidelity labels and real datasets with true noise. Additionally, contribution of our work includes study of effects of budget distribution among variable fidelity sources under different noise conditions and sensitivity analysis of the proposed model to its hyperparameters.
\section{Related work}
\label{sec:related_work}

A comprehensive introduction into GPs in the context of machine learning has been done previously \cite{Rasmussen2006gaussianprocesses}. We were guided by that book during the derivations of our algorithm. More detailed study \cite{nickisch2008approximations} of methods for approximate binary classification inference based on GPs demonstrates that Laplace Approximation is the fastest inference method with moderate accuracy, whereas Expectation Propagation is the most accurate, but runs approximately 10 times slower. The study outlines that the former should be considered when the error rate is the main metric, although the latter delivers more accurate class probabilities. In addition, when labels contain a lot of noise, the authors outline that all approximation methods tend to produce similar results.


Supervised classification in the presence of noise in labels \cite{natarajan2013learning} has been studied with class-conditional random Bernoulli noise, such classification problems have also got theoretical justification of their learnability.



Prior works extensively cover topics connected to multi-task learning \cite{zhang2017survey} in general and multi-output GPs \cite{alvarez2011computationally} in particular. Multi-fidelity regression based on GPs was also studied in a number of works \cite{kennedy2000predicting,le2013multi,le2014recursive}, including a co-kriging setup for fidelities with an exact inference schema for their regression \cite{forrester2007multi}. In our work, we adopt co-kriging for the classification problem by applying this setup on latent functions. Note that there is no exact inference schema for GP classification for single-fidelity case, nor for multi-fidelity one. 
Several recent works are dedicated to close problems, yet they all consider different aspects. For example, a work on the multivariate generalized linear geostatistical model with spatially structured bias \cite{doi:10.1111/rssa.12069} is close to ours, however, the model studied there doesn't take into account a scaling factor and the proposed inference is confined to MCMC method. A more recent work proposed a framework for handling heterogeneous outputs of GPs with stochastic variational inference \cite{moreno2018heterogeneous}, also there is a study of the application of heterogeneous multivariate GPs for joint species distribution modeling \cite{vanhatalo2018joint}. Compared to them our work is about a more specific model of multivariate GPs, that can be adapted to a classical algorithmic framework \cite{Rasmussen2006gaussianprocesses} without additional approximation techniques. This tailoring makes our method more robust to noise in labels and accurate than others that use more general models, as we show further in the experimental section. 

Heteroscedastic models \cite{goldberg1998regression,le2005heteroscedastic,kersting2007most} are complementary to our model in the sense that the former are about modeling input-dependent variance, whereas the latter is about modeling input-dependent bias between low- and high- fidelity processes.

There is a large branch of research on learning from multiple annotators \citep{zhang2016learning}, which partially intersects with the applications of our method. 
Early works in this direction started with different strategies of feature-agnostic labels integration and active learning for optimizing annotation costs \citep{sheng2008get,whitehill2009whose,DBLP:journals/corr/abs-1209-3686,ipeirotis2010quality}. A generative probabilistic model was proposed to estimate annotators expertise along with items annotation difficulty  \cite{NIPS2010_4074}, yet features are not observable for the model. Another work \cite{dekel2009vox} studied the problem of pruning low-quality annotators in order to improve the quality of the training set for binary classification problem. The same authors also built an algorithm on top of the SVM, that decreased influence of low-quality entries \citep{dekel2009good}. Several state-of-the-art works \citep{raykar2010learning,rodrigues2014gaussian,ruiz2016variational} model annotations as random Bernoulli labels dependent on the true class, which in turn is generated via latent Gaussian process; these works have similar setups and provide Variational Bayes and Expectation Propagation inferences for them. Overall, all these works deal with cases when many annotators are available, since otherwise their expertise (fidelity in our case) can barely be resolved. Moreover, \cite{arxiv1204.3511} showed that without a bit of gold-standard labels, that is, a high-fidelity source, crowd-sourcing labels integration methods will in some cases fail to resolve annotators expertise. Our work stands out from this branch of research due to explicitly fixing fidelities of data sources in our model, which takes into account each item's annotation difficulty. 

\section{Problem statement}
\label{sec:problem_statement}

There is a binary function $c:\Omega \rightarrow \{0, 1\}$ defined on the measurable set $\Omega \subset \mathbb{R}^d$. 
We have two samples:
\begin{equation}
\label{eq:samples}
D_H = \{(x^H_i, y^H_i)\}_{i=1}^{n_H} \textup{ and }
D_L = \{(x^L_i, y^L_i)\}_{i=1}^{n_L},
\end{equation}
where $x^L_i, x^H_i \in \Omega$ and $y^L_i, y^H_i \in \{0, 1\}$. Let us also denote $X_L = \{x^L_i\}_{i=1}^{n_L}$, and $X_H = \{x^H_i\}_{i=1}^{n_H}$. All notations are summarized in  Table \ref{tab:notations}.

Sample $D_H$ contains high-fidelity data, that is, it has much more reliable labels than $D_L$, which contains low-fidelity data respectively, so its labels can be biased and more noisy. Using the Bayesian approach we formally express this assumption with the following model: 
\begin{equation}
\label{eq:model}
\begin{split}
& c(x) = \mathbb{I}\left[f_H(x) > 0\right], \\
& p \left(y^H_i = 1|f_H(x^H_i)\right) = \sigma \left(f_H(x^H_i)\right), \\
& p \left(y^L_i = 1|f_L(x^L_i)\right) = \sigma \left(f_L(x^L_i)\right),
\end{split}
\end{equation}
where $\mathbb{I}$ is an indicator function; $\sigma(z) = \frac{1}{1 + exp(-z)}$ is a sigmoid function; $f_L$ and $f_H$ are Gaussian processes on $\Omega$. In our model we assume these processes are dependent via co-kriging model \citep{forrester2007multi}:
\begin{equation}
\label{eq:cokriging_dependency}
f_H(x^H_i) = \rho f_L(x^H_i) + \delta(x^H_i),
\end{equation}
where $\rho \in \mathbb{R}$ is a linear coefficient, and $\delta$ is a residual Gaussian process independent of $f_L$. Processes $f_L$ and $\delta$ have prior kernels $k_l$ and $k_d$ with hyper-parameters $\theta_l$ and $\theta_d$ respectively. Such dependency between latent processes has been on the one hand acknowledged in many engineering applications \cite{Forrester2008}, on the other hand, it corresponds to the optimal estimate of high-fidelity data according to the Theorem on normal correlation (see \cite{Liptser2001}, theorem 13.1). Parameter $\rho$ can reduce or increase the confidence of the high-fidelity model compared to the low-fidelity one, in particular, $\rho=1$ corresponds to the case when the low-fidelity source contains high-fidelity labels with additive noise. This parameter is also useful for cases, when low- and high- fidelity labels are mostly opposed to each other. Gaussian process $\delta$ can compensate predictions for input-dependent bias in low-fidelity data.

Finally, assuming models \eqref{eq:model} and \eqref{eq:cokriging_dependency} we would like to train a classifier $\hat{c}$ that estimates the function $c$ using samples \eqref{eq:samples}.

\section{Solution}
\label{sec:solution}

For simplicity of notation we omit specifying hyper-parameters ($\rho$ and parameters of kernels $\theta_l$, $\theta_d$) as conditions of probabilities in formulas below.

The predictive distribution of $f_H$ at $x_* \in \Omega$ is:
\begin{equation}
\label{MFGPC_predictive_gp}
p(f^H_*|D_L, D_H, x_*)
= \iint p(f^H_*|\mathbf{f}^L, \boldsymbol{\delta}, X_L, X_H, x_*) p(\mathbf{f}^L, \boldsymbol{\delta}|D_L, D_H) d \mathbf{f}^L d \boldsymbol{\delta},
\end{equation} 
where 
\begin{align*} 
\boldsymbol{\delta} &= \left(\delta(x^H_1), ..., \delta(x^H_{n_h})\right)\tran, \\ 
\mathbf{f}^L &= \left(f^L(x^L_1), ..., f^L(x^L_{n_L}), f^L(x^H_1), ..., f^L(x^H_{n_h})\right)\tran.
\end{align*}

The probability of $c$ to be $1$ at point $x_*$ can be expressed by marginalization of the predictive distribution:
\begin{equation}
\label{MFGPC_gp_class_prediction}
p(c(x_*) = 1|D_L, D_H, x_*) 
= \int \sigma(f^H_*) p(f^H_*|D_L, D_H, x_*) d f^H_*.
\end{equation}

Integrals \eqref{MFGPC_predictive_gp} and \eqref{MFGPC_gp_class_prediction} don't have analytic solutions, therefore they have to be numerically integrated or approximated analytically. 
In this work we use Laplace Approximation method to handle the former, whereas the predicted class label based on the latter integral can be easily calculated in the binary case once the predictive distribution is known or estimated \cite{bishop1995neural}:

\begin{equation*}
\begin{split}
&\hat{c}(x_*) = \mathbb{I}\left[\int \sigma(f^H_*) p(f^H_*|D_L, D_H, x_*) d f^H_*>\frac12\right] =\\
& = \mathbb{I}\left[\int f^H_* p(f^H_*|D_L, D_H, x_*) d f^H_*>0\right].
\end{split}
\end{equation*}

\subsection{Laplace Approximation}
Prediction based on GPs requires two steps \cite{Rasmussen2006gaussianprocesses}:
\begin{enumerate}
    \item Obtaining a latent predictive distribution for the test point via marginalizing the posterior distribution over all possible latent values at training points;
    \item Marginalizing it over all possible latent values at the test point in order to produce a probabilistic prediction.
\end{enumerate}

Unlike regression problem, where marginalizations are straightforward because all underlying components are Gaussian, prediction of classes is analytically intractable due to non-Gaussian likelihoods.

The idea of Laplace's method is to handle intractability at step 1 by applying a second order Taylor expansion of posterior's logarithm around its maximum. Thus we obtain a Gaussian approximation of the posterior distribution, which in turn makes approximate predictive distribution also Gaussian. Next, intractability of step 2 can be resolved by replacing marginalization with maximum a posteriori predictions \cite{bishop1995neural} or approximated with numerical techniques \cite{mackay1992evidence, williams1998bayesian}.

In the next three sections \ref{ss:mode-fitting}, \ref{ss:model_fitting} and \ref{ss:predictions} we will adjust our solution to fit the algorithmic framework for Laplace Approximation.
The key challenge in our case is dependence of $y^H_i$ on multiple latent components, which requires substantial modifications of basic algorithms.

\subsection{Mode-fitting}
\label{ss:mode-fitting}

The posterior distribution in integral \eqref{MFGPC_predictive_gp} is approximated with Gaussian distribution $q(\cdot)$: 

\begin{equation}
\label{eq:appox_q}
p(\mathbf{f}^L, \boldsymbol{\delta}|D_L, D_H) \approx q(\mathbf{f}^L, \boldsymbol{\delta}|D_L, D_H) 
= \Norm\left(\boldsymbol{\xi}=\begin{bmatrix}
\mathbf{f}^L\\
\boldsymbol{\delta}
\end{bmatrix}\Big|\ \boldsymbol{\hat\xi}, \Sigma\inv \right),
\end{equation}
where $\Sigma = -\nabla \nabla \log p(\boldsymbol{\xi}|D_L, D_H)\big|_{\boldsymbol{\xi}=\boldsymbol{\hat\xi}}$ and $\boldsymbol{\hat\xi} = \argmax{\boldsymbol{\xi}}\  p(\boldsymbol{\xi}|D_L, D_H)$. Thus, for obtaining approximate posterior distribution we need to calculate these parameters.

According to Bayes formula and monotonic increase of $\log$ function, the problem of finding $\boldsymbol{\hat\xi}$ is equivalent to:
\begin{equation}
\label{eq:laplace_argmax}
\argmax{\boldsymbol{\xi}}\  p(\boldsymbol{\xi}|D_L, D_H)  
= \argmax{\boldsymbol{\xi}}\ \left[\log p(\mathbf{y}^L, \mathbf{y}^H|\boldsymbol{\xi}) + \log p(\boldsymbol{\xi}|X_L, X_H)\right],
\end{equation}
where $\mathbf{y}^L = \left(y^L_1, ..., y^L_{n_L}\right)\tran$ and $\mathbf{y}^H = \left(y^H_1, ..., y^H_{n_H}\right)\tran$. Note that the probability of evidence is omitted, since it is independent of the argument. Problem \eqref{eq:laplace_argmax} has a unique solution, see details in the \ref{sec:correctness}. 

Let us now define $\Psi(\boldsymbol{\xi}) \overset{\mathsmaller{\triangle}}{=} \log p(\mathbf{y}^L, \mathbf{y}^H|\boldsymbol{\xi}) + \log p(\boldsymbol{\xi}|X_L, X_H),$ and look at its components in more detail. Let also $X = X_L \cup X_H$.

The prior distribution of $\boldsymbol{\xi}$ is normal:
\begin{equation}
\label{eq:prior_xi}
p(\boldsymbol{\xi}|X_L, X_H) \sim \Norm\left(0, K = \begin{bmatrix}
k_l(X, X) & 0\\
0 & k_d(X_H, X_H)
\end{bmatrix}\right).
\end{equation}

Log-likelihood is:
\begin{equation*}
\lambda \eqdef \log p(\mathbf{y}^L, \mathbf{y}^H|\boldsymbol{\xi}) =
\underset{i = 1..n_l}\sum{\log\ \sigma\big(\widetilde{y}^L_if^L(x^L_i)\big)} 
+ \underset{i = 1..n_h}\sum{\log\ \sigma\big(\widetilde{y}^H_i(\rho f^L(x^H_i) + \delta_i)\big)},
\end{equation*}
where for simplicity of notation we use: \[\delta_i = \delta(x_i^H)\textup{, }\widetilde{y}^L_i = (2y^L_i - 1)\textup{, and }\widetilde{y}^H_i = (2y^H_i - 1).\] 

Having figured out expressions for components of  $\Psi(\boldsymbol{\xi})$, the solution of problem \eqref{eq:laplace_argmax} can be found with iterative Newton's method:
\begin{equation*}
    \boldsymbol{\hat\xi}^{\textup{ new}} = \boldsymbol{\hat\xi}^{\textup{ old}} - (\nabla\nabla\Psi)\inv \nabla\Psi\big|_{\boldsymbol{\xi}=\boldsymbol{\hat\xi}^{\textup{ old}}}.
\end{equation*}

\subsection{Model selection}
\label{ss:model_fitting}
Let us denote $\widetilde{q}(.)$ a Gaussian approximation of the marginal likelihood $p(\mathbf{y}^L, \mathbf{y}^H|X_L, X_H, \rho, \theta_l, \theta_d)$. Model selection implies finding hyper-parameters $\rho$, $\theta_l$, and $\theta_d$ that maximize the approximate log marginal likelihood (this approximation is obtained similarly to the single-fidelity case \cite{Rasmussen2006gaussianprocesses}):
\begin{equation}
\label{eq:log_marginal_likelihood}
\mathcal{L} \eqdef \log \widetilde{q}(\mathbf{y}^L, \mathbf{y}^H|X_L, X_H, \rho, \theta_l, \theta_d)
= -\frac12 \boldsymbol{\hat\xi}\tran K\inv \boldsymbol{\hat\xi} + \lambda - \frac12\log|B|,
\end{equation}
where $B = I + W^\frac12 K W^\frac12$ and 
\begin{equation}
\label{eq:W}
 W \overset{\mathsmaller{\triangle}}{=} -\nabla\nabla_{\boldsymbol{\xi}} \log p(\mathbf{y}^L, \mathbf{y}^H|\boldsymbol{\xi})
= \begin{bmatrix}
A & 0 & 0 \\
0 & \rho^2 D & \rho D \\
0 & \rho D & D
\end{bmatrix};
\end{equation}\\
\begin{align*}
    &A = \nabla\nabla_{f^L(X_L)}\lambda = \text{diag}\left(\omega \left(f^L(x^L_i) \right)\big|_{i=1..n_l}\right), \\
    &D = \nabla\nabla_{\boldsymbol{\delta}}\lambda = \text{diag}\left(\omega \left(\rho f^L(x^H_i) + \delta_i \right)\big|_{i=1..n_h}\right), \\
    &\omega(z) = \sigma'(z) = \sigma(z)(1 - \sigma(z)).
\end{align*}

Unlike single-fidelity case, $W$ in multi-fidelity case is non-diagonal, so computation of its square root is not straightforward. We have derived the exact formula for its fast and numerically stable calculation:
\begin{equation}
W^\frac12 = \begin{bmatrix}
A^{\frac{1}{2}} & 0  \\
0 & \frac{1}{\sqrt{\rho^2 + 1}}\begin{bmatrix}
\rho^2 & \rho\\
\rho & 1
\end{bmatrix}\otimes D^{\frac{1}{2}}
\end{bmatrix},
\end{equation}
note that matrices $A$ and $D$ are diagonal, so their square roots are easily calculated.

In order to maximize log marginal likelihood \eqref{eq:log_marginal_likelihood}, one can use gradient-based optimization, which requires its partial derivatives w.r.t. hyper-parameters.

Derivatives of $\mathcal{L}$ and $\boldsymbol{\hat \xi}$ w.r.t. kernel hyper-parameters $\theta_l$ and $\theta_d$ are analogous to formulas in the single-fidelity case (\cite{Rasmussen2006gaussianprocesses}, section 5.5.1), thus we omit them here, except the formula 5.23 from \cite{Rasmussen2006gaussianprocesses} for partial derivatives of $\mathcal{L}$ w.r.t. components of $\boldsymbol{\hat \xi}$, which reduces calculation of trace to multiplication of $i$-th diagonal elements. That reduction doesn't take place for multi-fidelity case, since $\frac{\partial W}{\partial \hat\xi_i}$ is not diagonal in general. We propose the following modification of that formula:
\begin{equation}
\label{eq:d_L_d_xi}
\frac{\partial \mathcal{L}}{\partial \hat\xi_i} = -\frac{1}{2}\textup{tr}\left((K\inv + W)\inv \frac{\partial W}{\partial \hat\xi_i}\right) 
= -\frac{1}{2}\underset{\textup{all elements}}{\sum}\left((K\inv + W)\inv \circ \frac{\partial W}{\partial \hat\xi_i}\right),
\end{equation}
where $\circ$ is an Hadamard (entrywise) product. Note that $\frac{\partial W}{\partial \hat\xi_i}$ is a sparse matrix that has at most 4 non-zero elements (see details in the \ref{seq:components_of_d_L_d_xi}), therefore computation time of the derivatives remains linear.

Derivative of $\mathcal{L}$ w.r.t to $\rho$ is:
\begin{equation}
\label{eq:d_L_d_rho}
\frac{\partial \mathcal{L}}{\partial \rho} = -\boldsymbol{\hat\xi}\tran K\inv \frac{\partial \boldsymbol{\hat\xi}}{\partial \rho} + \frac{\partial \lambda}{\partial \rho} - \frac{1}{2}\frac{\partial \log |B|}{\partial \rho}.
\end{equation}
Note that in our setup $K$ doesn't depend on $\rho$.
Now let's look into components of \eqref{eq:d_L_d_rho} in more detail. 
We differentiate by $\rho$ the necessary condition of the maximum $\nabla \Psi(\boldsymbol{\xi})|_{\boldsymbol{\xi} = \boldsymbol{\hat \xi}} = 0$, where $\nabla \Psi(\boldsymbol{\xi}) = \nabla_{\boldsymbol{\xi}} \lambda - K\inv \boldsymbol{\xi}$, obtaining an equation on $\boldsymbol{\xi}$:
\begin{equation}
\label{eq:d_xi_d_rho}
\begin{split}
& \frac{\partial \boldsymbol{\hat\xi}}{\partial \rho} = K \left(-W\frac{\partial \boldsymbol{\hat \xi}}{\partial \rho}  + \left.\frac{\partial \nabla_{\boldsymbol{\xi}}\lambda\big|_{\boldsymbol{\xi} = \boldsymbol{\hat\xi}}}{\partial \rho}\right|_{\textup{explicit}}\right) \Rightarrow \\
& \Rightarrow \frac{\partial \boldsymbol{\hat\xi}}{\partial \rho} = (I + K W)\inv K \left( \left. \frac{\partial \nabla_{\boldsymbol{\xi}}\lambda\big|_{\boldsymbol{\xi} = \boldsymbol{\hat\xi}}}{\partial \rho}\right|_{\textup{explicit}} \right),
\end{split}
\end{equation}
where the components of the \emph{explicit} term in formula \eqref{eq:d_xi_d_rho} and derivatives of $\lambda$ w.r.t. components of $\boldsymbol{\xi}$ are provided in Table \ref{tb:explicit_derivative}.

\begin{table*}[t]
\centering
\caption{Components of $\boldsymbol{\xi}$, corresponding derivatives of $\lambda$ and the explicit term in \eqref{eq:d_xi_d_rho}; here $f^L_i = f^L(x^L_i)$ and $f^H_i = \rho f^L(x^H_{i}) + \delta(x^H_{i}) $.} 
\label{tb:explicit_derivative}

\begin{center}
\begin{tabular}{ r|c c }
\hline
components of $\boldsymbol{\xi}$ & components of  $\nabla_{\boldsymbol{\xi}} \lambda$ & components of $\frac{\partial \nabla_{\boldsymbol{\xi}}\lambda|_{\boldsymbol{\xi} = \boldsymbol{\hat\xi}}}{\partial \rho}\big|_{\textup{explicit}}$\\
\hline

$f^L(X_L)$  & $y^L_i - \sigma(f^L_i)$ & 0 \\

$f^L(X_H)$ &$\rho (y^H_{i} - \sigma(f^H_i)) $ & $y^H_{i} - \sigma(f^H_i) - \rho f^L(x^H_{i}) \omega(f^H_i)$ \\

$\delta(X_H)$ & $y^H_{i} - \sigma(f^H_i)$ & $-f^L(x^H_{i}) \omega(f^H_i)$ \\

\hline
\end{tabular}
\end{center}
\end{table*}

Next component of \eqref{eq:d_L_d_rho} is:
\begin{equation}
 \frac{\partial \lambda}{\partial \rho} 
= \sum_{i=1..n_h}\widetilde y_i^H f^L(x_i^H) \left(1 - \sigma\left(\widetilde y_i^H (\rho f^L(x_i^H) + \delta(x_i^H))\right)\right) 
+ \sum_i \frac{\partial \lambda}{\partial \xi_i} \frac{\partial \xi_i}{\partial \rho}.
\end{equation}

The last component is (see the derivation in the \ref{seq:inference_d_log_det_B_d_rho}):
\begin{equation}
\label{eq:d_log_det_B_d_rho}
\frac{\partial \log |B|}{\partial \rho} = \sum_{\textup{all elements}} \left((K\inv + W)\inv \circ \frac{\partial  W}{\partial \rho}\right),
\end{equation}
where 
\begin{align*}
    &\frac{\partial W}{\partial \rho} = \begin{bmatrix}
    0 & 0  \\
    0 & \begin{bmatrix}
    \rho^2 & \rho\\
    \rho & 1
    \end{bmatrix}\otimes \frac{\partial D}{\partial \rho}\Big|_{\text{explicit}} +  \begin{bmatrix}
    2\rho & 1\\
    1 & 0
    \end{bmatrix}\otimes D
    \end{bmatrix} 
    +\sum_i \frac{\partial W}{\partial \xi_i} \frac{\partial \xi_i}{\partial \rho};\\
    &\frac{\partial D}{\partial \rho}\Big|_{\text{explicit}} = \text{diag}\left(f^L(x^H_i)\zeta(f^H_i) \big|_{i=1..n_h}\right)\textup{ and }\zeta(x) = \sigma''(x).
\end{align*}


\subsection{Predictions}
\label{ss:predictions}
Once we know estimates of parameters and hyper-parameters, we can use an ordinary schema of exact multi-fidelity posterior from \cite{forrester2007multi} to obtain \emph{MAP predictions}:
\begin{equation}
\mathbb{E}[f_*|D_L, D_H, x_*] \approx \mathbb{E}_q[f_*|D_L, D_H, x_*] = \widetilde k\tran_* \widetilde K\inv \mathbf{\hat f},
\end{equation}
where 
\begin{align*}
    &\widetilde k\tran_* = \begin{bmatrix}k_l(x_*, X_L) & \rho k_l(x_*, X_H) + k_d(x_*, X_H)\end{bmatrix}, \\
    &\boldsymbol{\hat \xi} = \begin{bmatrix}\hat f^L(X_L) & \hat f^L(X_H) & \hat \delta(X_H)\end{bmatrix}\tran, \\
    &\widetilde K = \begin{bmatrix}k_l(X_L, X_L) & \rho k_l(X_L, X_H) \\
     \rho k_l(X_H, X_L) & \rho^2 k_l(X_H, X_H) + k_d(X_H, X_H) \end{bmatrix}, \\
    &\mathbf{\hat f} = \begin{bmatrix}\hat f^L(X_L) \\ \rho \hat f^L(X_H) + \hat \delta(X_H)\end{bmatrix}.
\end{align*}

\section{Experiments}
\label{sec:experiments}

We compared our model with a number of baseline approaches. 
The baselines are built upon ordinary Gaussian Process Classifier (\textbf{\gpc{}}), Logistic Regression (\textbf{\logit{}}) and Gradient Boosting Classifier (\textbf{\xgb{}}). We trained those baselines in three modes: 
\begin{enumerate}
    \item Training only on high-fidelity data (no prefix);
    \item Training on concatenated high- and low- fidelity data (with prefix \textbf{\concMF{}});
    \item Stacking low-fidelity predictions, that is, predictions of a classifier trained on low-fidelity data were used as additional features for training the classifier on high-fidelity data (with prefix \textbf{\stackedMF{}}).
\end{enumerate}
All GPs-based methods used isotropic RBF kernel. More details regarding the experimental implementation see in the \ref{seq:specifications_of_implementation}.

\subsection{Evaluation metrics}
In order to compare performance of various methods we use areas under receiver operating characteristic curves \cite{hanley1982meaning} (\ROCAUC{}) metric. 

Further, to aggregate performance across many tests and datasets, we average \ROCAUC{} over them. We also  supplement results with figures of \emph{\ROCAUC{} profiles}, which show the share of tests where the corresponding methods had greater \ROCAUC{} than the threshold pointed on the abscissa axis. The rule of thumb for assessing such profiles is the higher the curve, the better the corresponding method.

\subsection{Datasets}

We evaluated models on three groups of datasets: 
\begin{enumerate}
    \item Artificial datasets\footnote{We published them in this repository  \texttt{https://github.com/user525/mfgpc}}: we constructed datasets by virtue of the model \eqref{eq:model} and \eqref{eq:cokriging_dependency}. Latent functions $f_L$ and $\delta$ were generated as instances of Gaussian processes, linear coefficients $\rho$ were adjusted to the desired  discrepancy (noise level) between low- and high- fidelities. We used input dimensions 2, 5, 10, and 20. For each of them, we generated 10 datasets.
    \item Datasets from Penn Machine Learning Benchmarks repository \citep{Olson2017PMLB}: we selected several representative benchmarks with different types of features, namely \texttt{diabetes} (\diabetes{}), \texttt{german} (\german{}), \texttt{waveform}-40 (\waveform{}), \texttt{satimage} (\satimage{}), \texttt{splice} (\splice{}), \texttt{spambase} (\spambase{}), \texttt{hypothyroid} (\hypothyroid{}), and \texttt{mushroom} (\mushroom{}). Since some datasets had multiple classes, we also selected one target representative class to test its classification against others: class 0 for \texttt{waveform-40} and \texttt{splice}, class 1 for \texttt{satimage} and class 2 for \texttt{diabetes}. Low-fidelity labels were generated by flipping original labels with the specified probability (noise level).
    \item Real datasets: we used \texttt{music\_genre} (\musicgenre{}) and \texttt{sentiment\_polarity} (\sentimentpolarity{}) from \cite{rodrigues2014gaussian}, which had been annotated with crowd-sourcing. Each object in those datasets was labeled by multiple annotators, therefore we considered majority voting statistic over object labels as high-fidelity and a single random annotation as low-fidelity. Such an approach to model fidelity is reasonable in the context of crowd-sourced annotations where each of them costs some amount of resources (e.g. money or time). For example, some objects are easy to classify with machine learning algorithms, thus one vote would be enough to annotate them, whereas for complex objects many votes are necessary for obtaining good confidence in labels. Finally, since \texttt{music\_genre} dataset had multiple classes, we tested each of them with the one-vs-all scheme as separate datasets. 
\end{enumerate}

\subsection{Comparison of methods}

For datasets in groups 1 and 2 we compared methods with the state-of-the-art \hetmogp{} \cite{moreno2018heterogeneous}. 
For datasets on crowd-sourcing annotation (group 3) we also compared our method with the state-of-the-art method \gpma{} \citep{rodrigues2014gaussian}. No comparison was made with the method of \cite{ruiz2016variational}, since we couldn't find publicly available source code.

At the outset, we verified our implementation of Laplace inference by comparing its predictions with those of MCMC (implemented with PyMC3 package \cite{salvatier2016probabilistic}) with the same hyper-parameters on real datasets from group 2 ensuring that true posteriors are non-Gaussian. Each training set contained 75 randomly sampled high fidelity observations and flip probability 0.2 in low-fidelity observations. The typical results of comparison are shown in Figure \ref{fig:mcmc_vs_lapl} and Table \ref{tab:mcmc_vs_lapl}. The overall performance of two inference approaches is on par, whereas correlation behavior resembles patterns observed in single-fidelity GPs classification (\cite{nickisch2008approximations}, figure~6), which lends evidence supporting the correctness of our method.

\begin{figure*}[t]
    \centering
    \begin{subfigure}[t]{0.22\textwidth}

        \centering
        \begin{tikzpicture}
          \node[inner sep=0pt] (A) {\includegraphics[width=\textwidth]{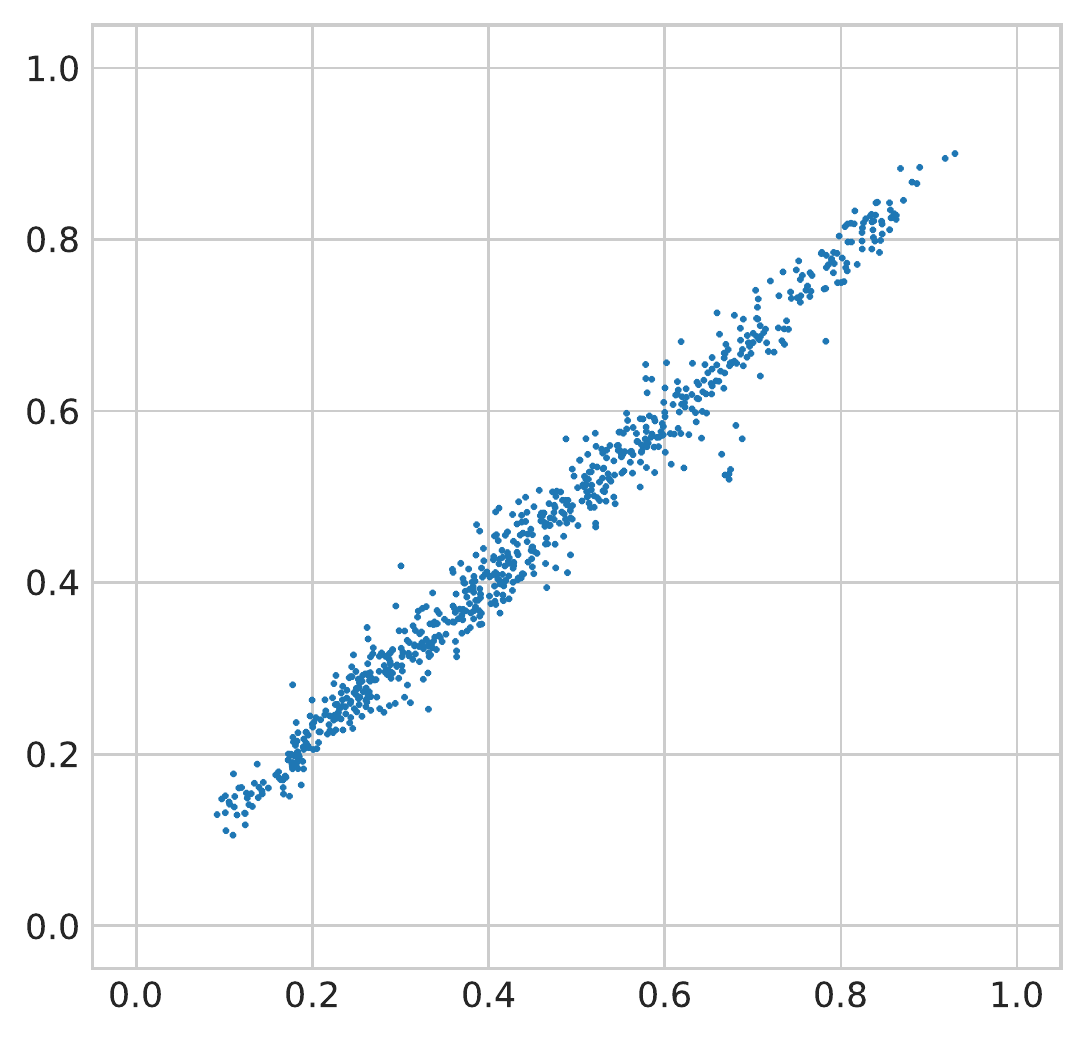}};
          \node (B) at ($(A.south)!-.03!(A.north)$) {\tiny MCMC};
          \node[rotate=90] (C) at ($(A.west)!-.03!(A.east)$) {\tiny Laplace};
        \end{tikzpicture}
        \caption{\diabetes{}}
        \label{subfig:mcmc_vs_lapl_a}
    \end{subfigure}%
    \hfill
    \begin{subfigure}[t]{0.22\textwidth}
        \centering
        \begin{tikzpicture}
          \node[inner sep=0pt] (A) {\includegraphics[width=\textwidth]{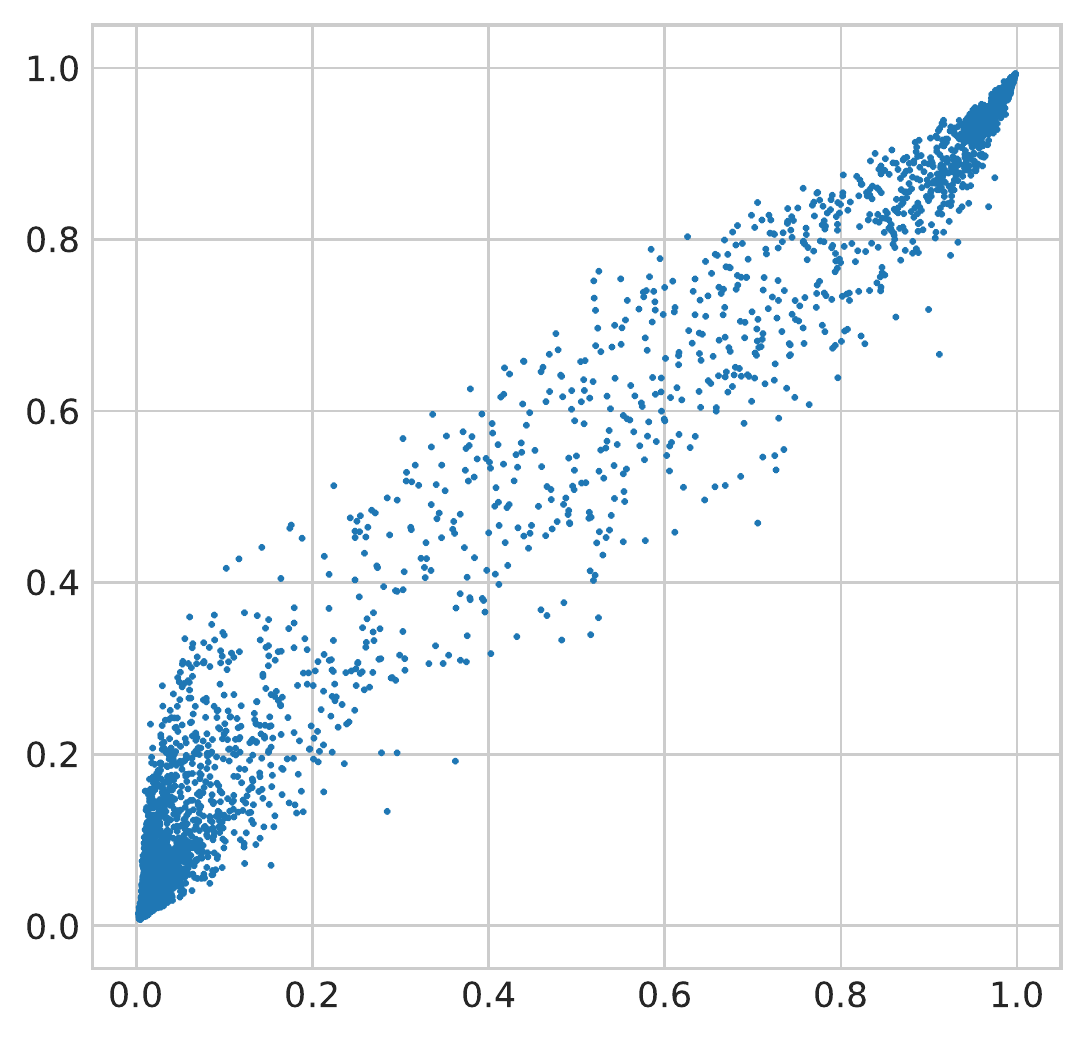}};
          \node (B) at ($(A.south)!-.03!(A.north)$) {\tiny MCMC };
          \node[rotate=90] (C) at ($(A.west)!-.03!(A.east)$) {\tiny Laplace};
        \end{tikzpicture}
        \caption{\satimage{}}
        \label{subfig:mcmc_vs_lapl_b}
    \end{subfigure}
    \hfill
    \begin{subfigure}[t]{0.22\textwidth}
        \centering
        \begin{tikzpicture}
          \node[inner sep=0pt] (A) {\includegraphics[width=\textwidth]{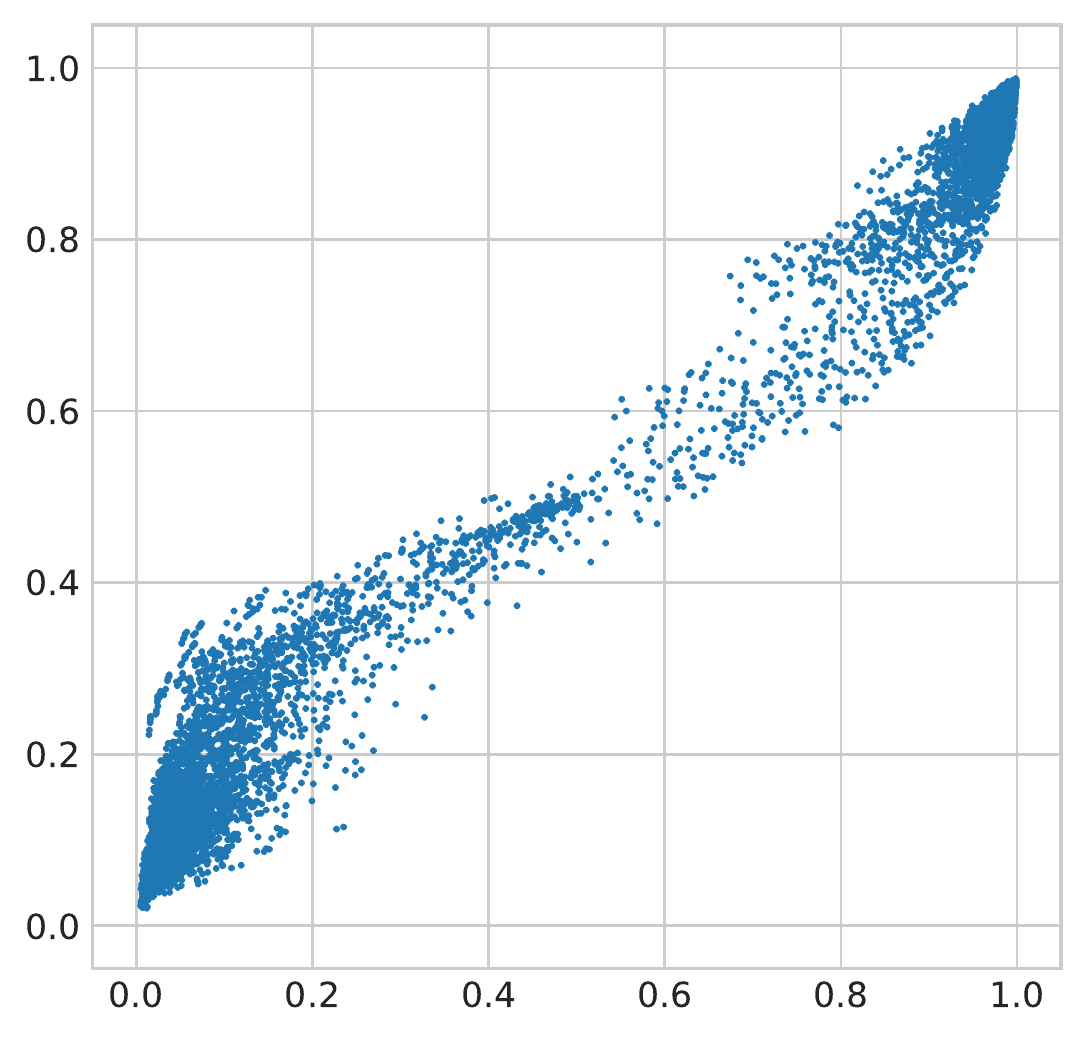}};
          \node (B) at ($(A.south)!-.03!(A.north)$) {\tiny MCMC };
          \node[rotate=90] (C) at ($(A.west)!-.03!(A.east)$) {\tiny Laplace};
        \end{tikzpicture}
        \caption{\mushroom{}}
        \label{subfig:mcmc_vs_lapl_c}
    \end{subfigure}
    \hfill
    \begin{subfigure}[t]{0.22\textwidth}
    
        \centering
        \begin{tikzpicture}
          \node[inner sep=0pt] (A) {\includegraphics[width=\textwidth]{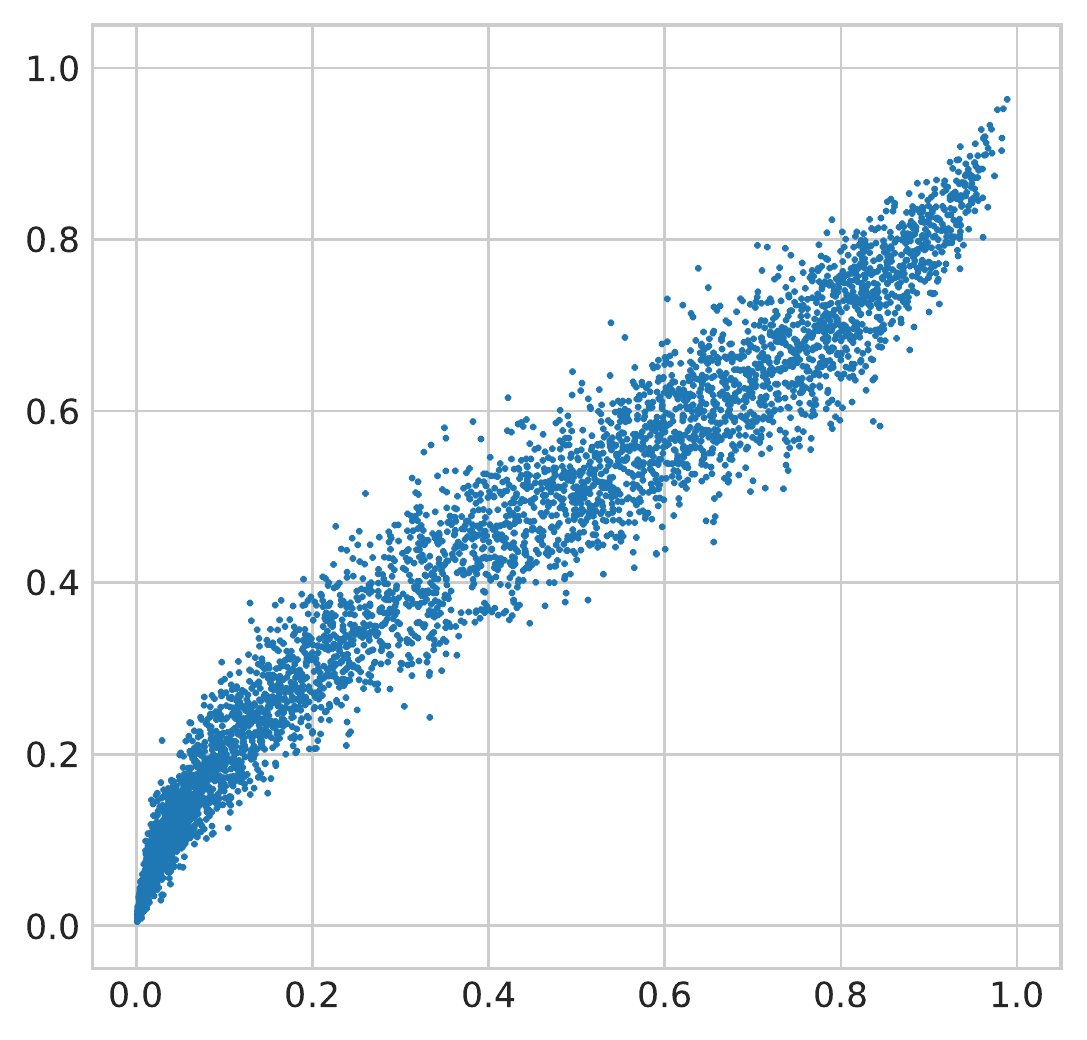}};
          \node (B) at ($(A.south)!-.03!(A.north)$) {\tiny MCMC };
          \node[rotate=90] (C) at ($(A.west)!-.03!(A.east)$) {\tiny Laplace};
        \end{tikzpicture}
        \caption{\waveform{}}
        \label{subfig:mcmc_vs_lapl_d}
    \end{subfigure}
    \caption{Comparison of predicted class probabilities with multi-fidelity MCMC and Laplace inference on datasets from group 2: typical cases of correlations.}
    \label{fig:mcmc_vs_lapl}
\end{figure*}
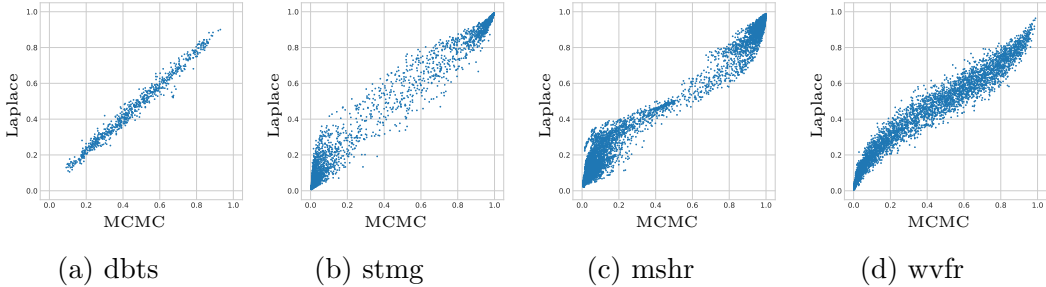
\begin{table*}[t]
    \centering
    \caption{Comparison of \ROCAUC{} in a single run for MCMC and Laplace inference on datasets from group 2 during verification tests.}
    \label{tab:mcmc_vs_lapl}
    
    \begin{adjustbox}{center}
    \begin{tabular}{l|ccccccccc}
    \hline
     & \diabetes{}  & \german{}  & \satimage{}  & \mushroom{}  & \splice{}  & \spambase{}  & \hypothyroid{}  & \waveform{} \\
    \hline
    \mfgpc{} Laplace & 0.815 & 0.787 & 0.998 & 0.999 & 0.940 & 0.942 & 0.614 & 0.932 \\
    \mfgpc{} MCMC & 0.809 & 0.780 & 0.997 & 0.999 & 0.922 & 0.946 & 0.624 & 0.927 \\
    \hline
    \end{tabular}
    \end{adjustbox}
    
\end{table*}

The main evaluation procedure was the following: for a single test, we selected a small random subsample of high fidelity observations and 3 times larger subsample of low fidelity observations. We trained all methods on those subsamples and evaluated predictions on the high-fidelity test set. For each dataset, we run 3 tests with different random subsamples, except \texttt{sentiment\_polarity}, for which we run 15 tests.

We report average \ROCAUC{} across all tests and methods in figures \ref{fig:artifitial_ROCAUC}, \ref{fig:results_group_2_noise_0_2}, \ref{fig:results_group_2_noise_0_4} (see also appendix with corresponding tables \ref{tab:results_group_1}, \ref{tab:results_group_2_noise_0_2}, \ref{tab:results_group_2_noise_0_4}) and table \ref{tab:results_group_3}. For those tests, each training set contained 75 high fidelity observations. Methods that performed not worse than 1 percent compared to the best result on the dataset are highlighted with bold.

\begin{figure*}[t]
    \centering
    
    \begin{subfigure}[t]{0.30\textwidth}
        \centering
        \begin{tikzpicture}
          \node[inner sep=0pt] (A) {\includegraphics[width=\textwidth]{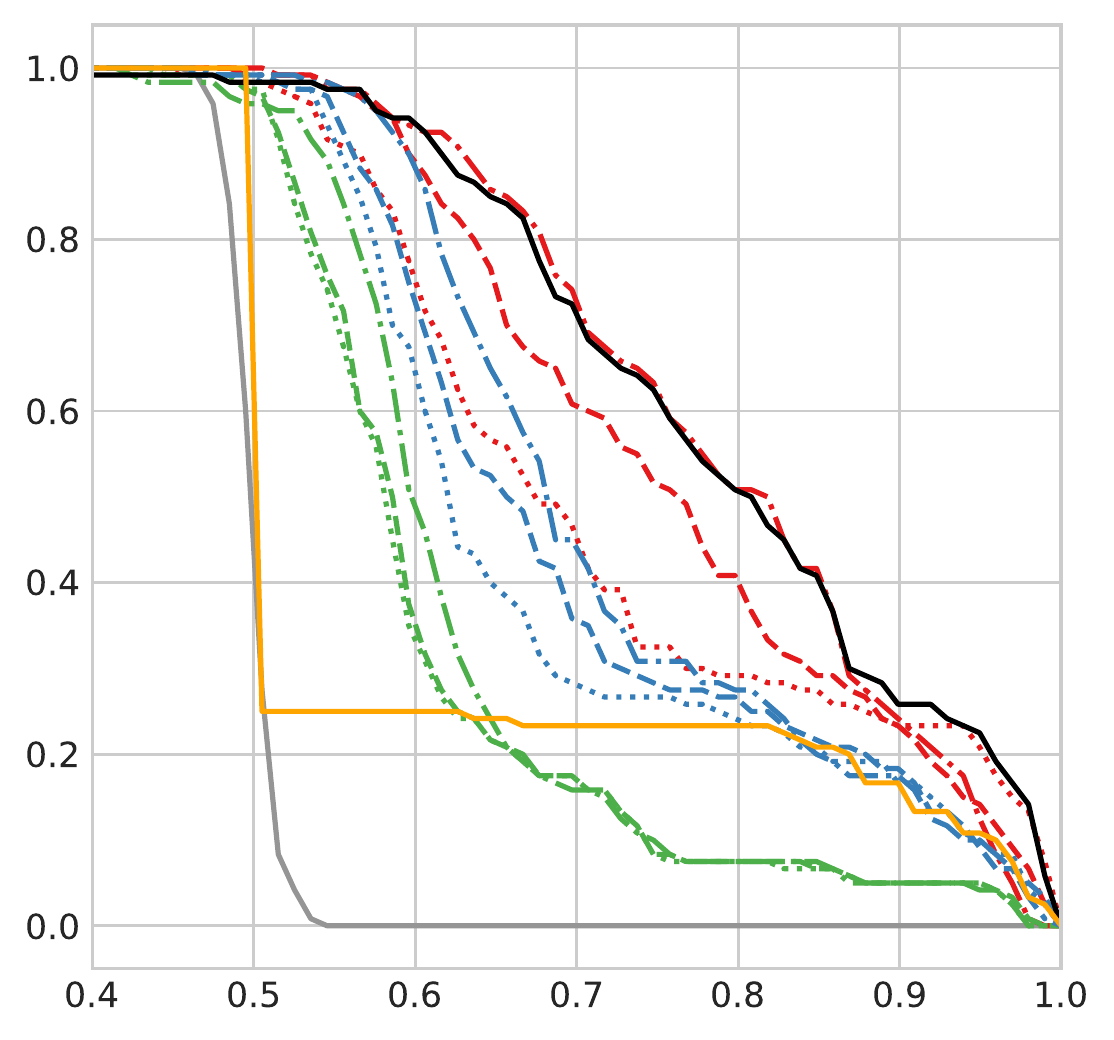}};
          \node (B) at ($(A.south)!-.03!(A.north)$) {\small Threshold };
          \node[rotate=90] (C) at ($(A.west)!-.03!(A.east)$) {\small Share};
        \end{tikzpicture}
        
        \caption{Noise level 0.2}
    \end{subfigure}%
    \hspace{1cm}
    \begin{subfigure}[t]{0.30\textwidth}
        \centering
        \begin{tikzpicture}
          \node[inner sep=0pt] (A) {\includegraphics[width=\textwidth]{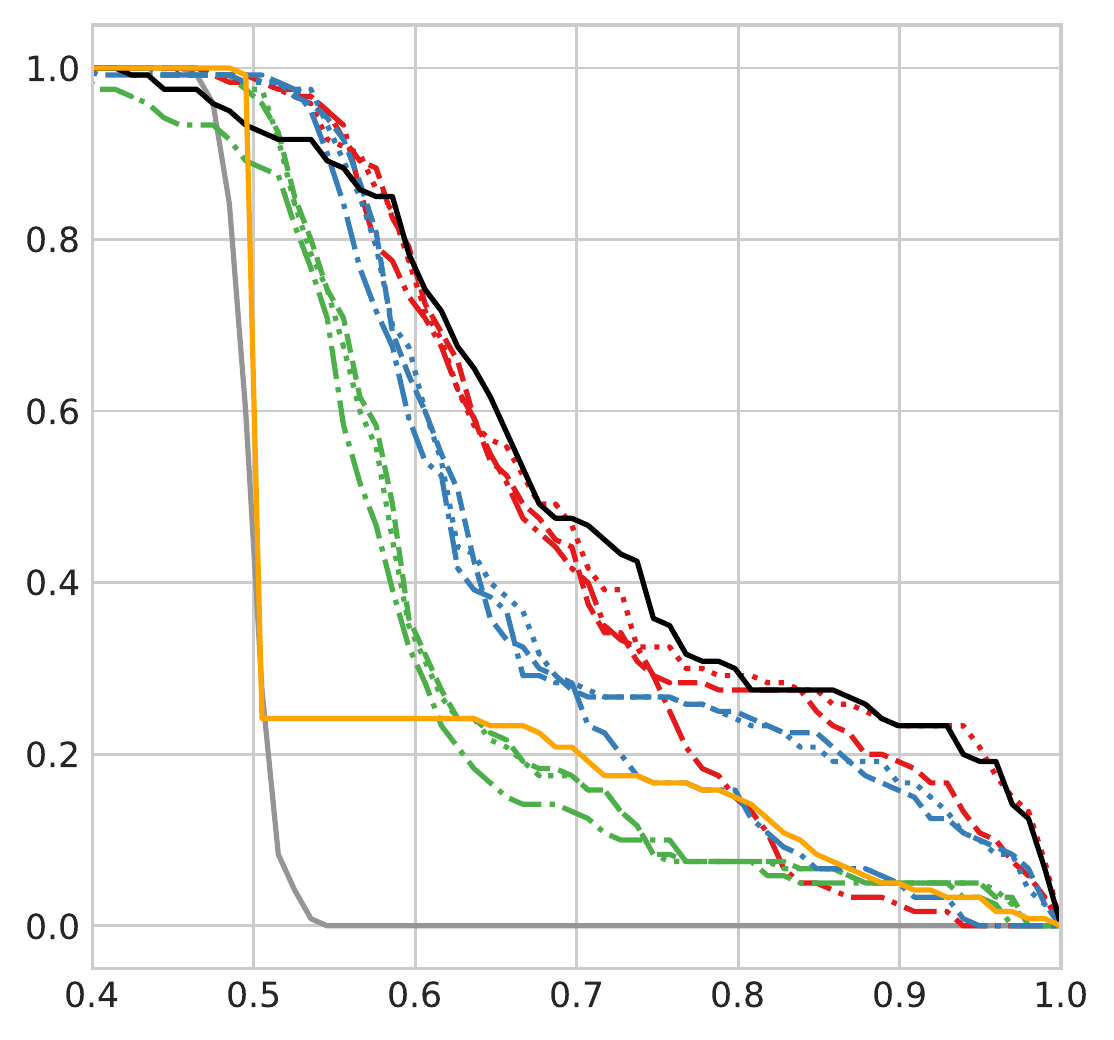}};
          \node (B) at ($(A.south)!-.03!(A.north)$) {\small Threshold };
          \node[rotate=90] (C) at ($(A.west)!-.03!(A.east)$) {\small Share};
        \end{tikzpicture}
        
        \caption{Noise level 0.4}
    \end{subfigure}
    \hspace{1cm}
    \begin{minipage}{0.155\textwidth}
        \vspace{-100pt}
        \small
        \begin{adjustbox}{width=1\textwidth}
          \begin{tabular}[t]{|l|}
          
            \hline
            \plainline{mfgpc} \mfgpc{} \\
            \plainline{major_vote} random \\
            \dottedline{gpc} \gpc{} \\
            \dottedline{logit} \logit{} \\
            \dottedline{xgb} \xgb{}\\
            \dashdottedline{conc_mf_gpc_x3} \concMF{} \gpc{} \\
            \dashdottedline{conc_mf_logit_x3} \concMF{} \logit{} \\
            \dashdottedline{conc_mf_xgb_x3} \concMF{} \xgb{} \\
            \dashedline{stacked_mf_gpc_x3} \stackedMF{} \gpc{} \\
            \dashedline{stacked_mf_logit_x3} \stackedMF{} \logit{} \\
            \dashedline{stacked_mf_xgb_x3} \stackedMF{} \xgb{} \\
            \plainline{gpma} \gpma{} \\
            \plainline{hetmogp} \hetmogp{} \\
            \hline
            \end{tabular}
        \end{adjustbox}
    \end{minipage}
    \caption{\ROCAUC{} profiles for artificial datasets from group 1.}
    \label{fig:artifitial_profiles}
    
\end{figure*}

\begin{figure*}[t]
    \centering
    \begin{subfigure}[t]{0.22\textwidth}
        \centering
        \begin{tikzpicture}
          \node[inner sep=0pt] (A) {\includegraphics[width=\textwidth]{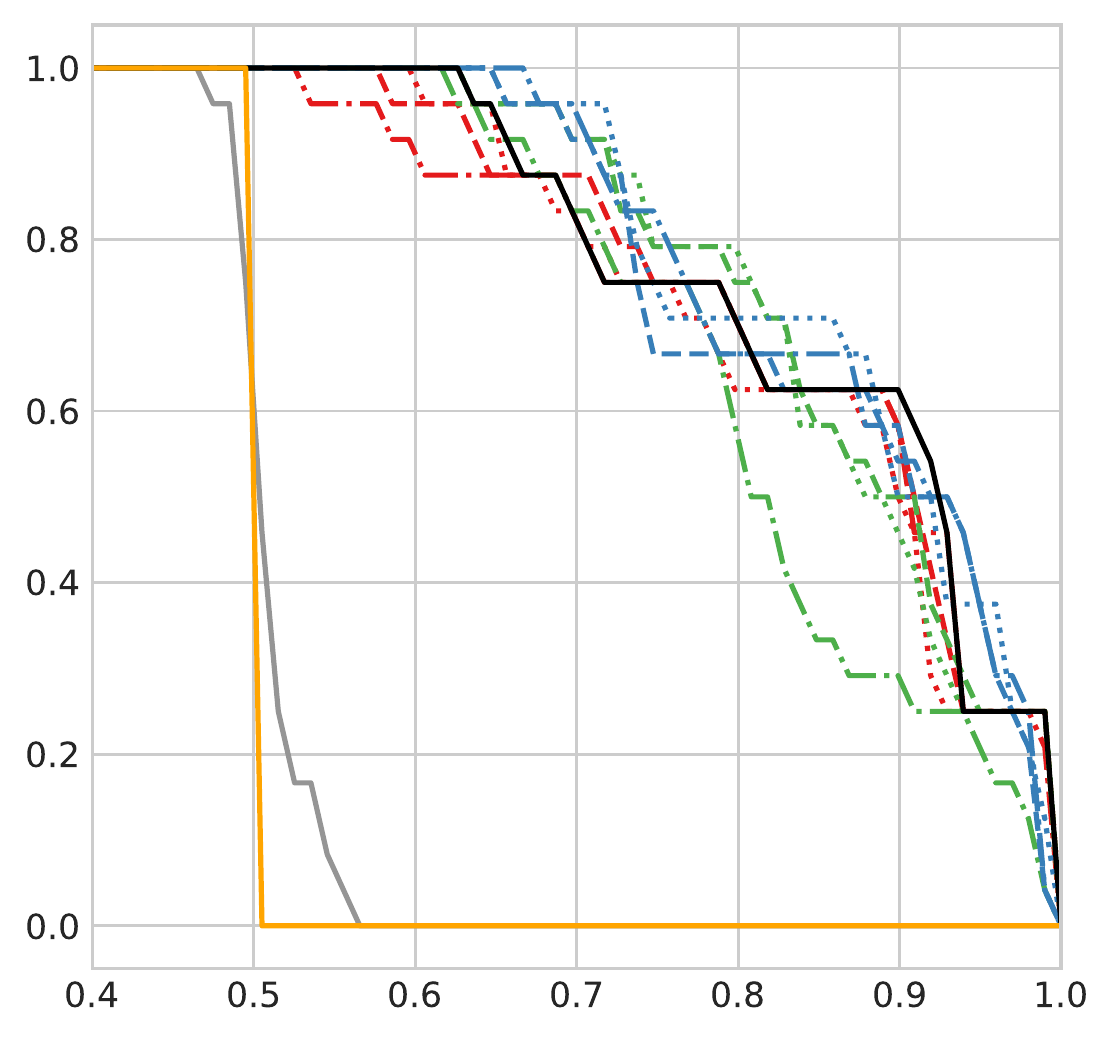}};
          \node (B) at ($(A.south)!-.03!(A.north)$) {\tiny Threshold };
          \node[rotate=90] (C) at ($(A.west)!-.03!(A.east)$) {\tiny Share};
        \end{tikzpicture}
        \caption{\shortstack{Group 2\\noise level 0.2}}
    \end{subfigure}%
    \hfill
    \begin{subfigure}[t]{0.22\textwidth}
        \centering
        \begin{tikzpicture}
          \node[inner sep=0pt] (A) {\includegraphics[width=\textwidth]{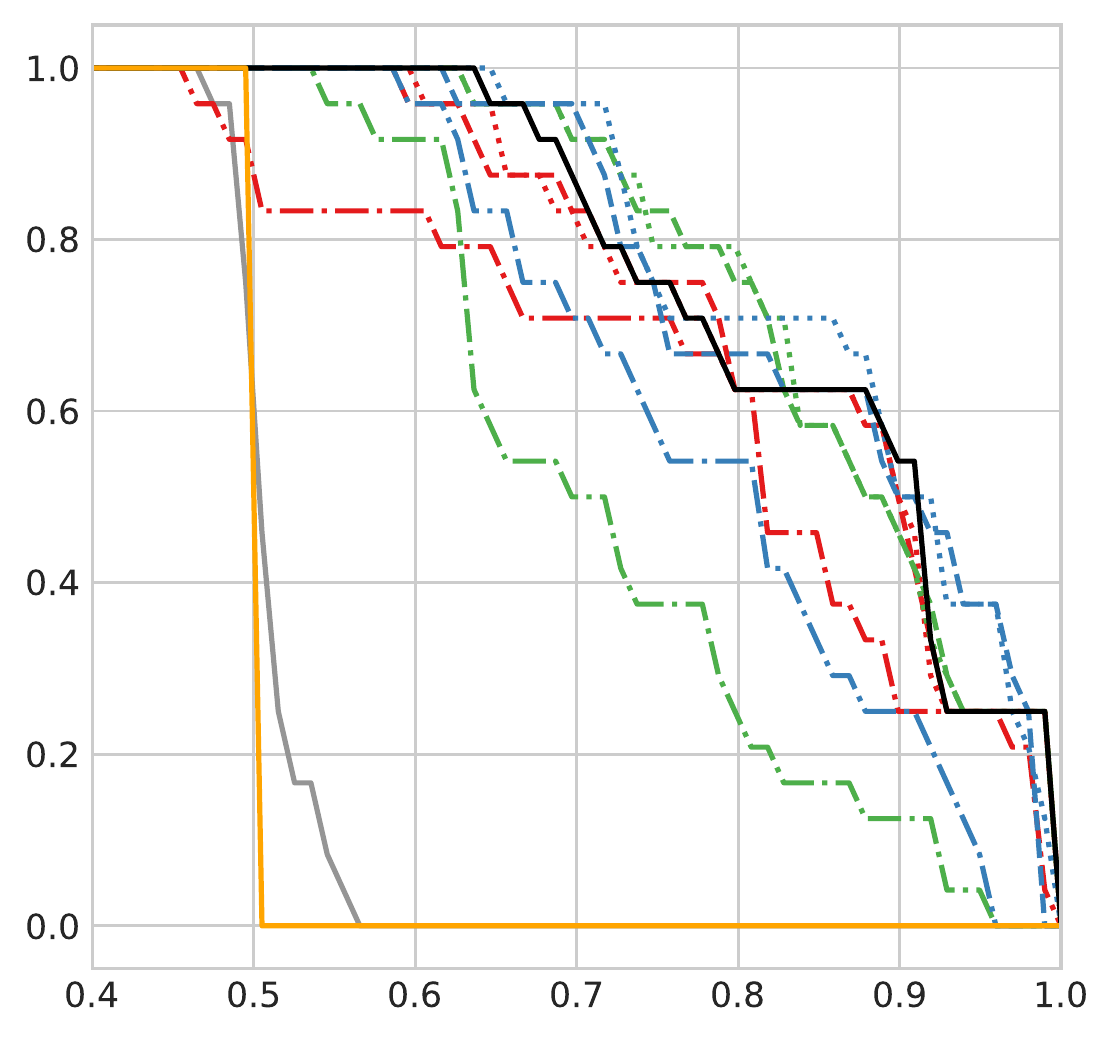}};
          \node (B) at ($(A.south)!-.03!(A.north)$) {\tiny Threshold };
          \node[rotate=90] (C) at ($(A.west)!-.03!(A.east)$) {\tiny Share};
        \end{tikzpicture}
        
        \caption{\shortstack{Group 2\\noise level 0.4}}
    \end{subfigure}
    \hfill
    \begin{subfigure}[t]{0.22\textwidth}
        \centering
        \begin{tikzpicture}
          \node[inner sep=0pt] (A) {\includegraphics[width=\textwidth]{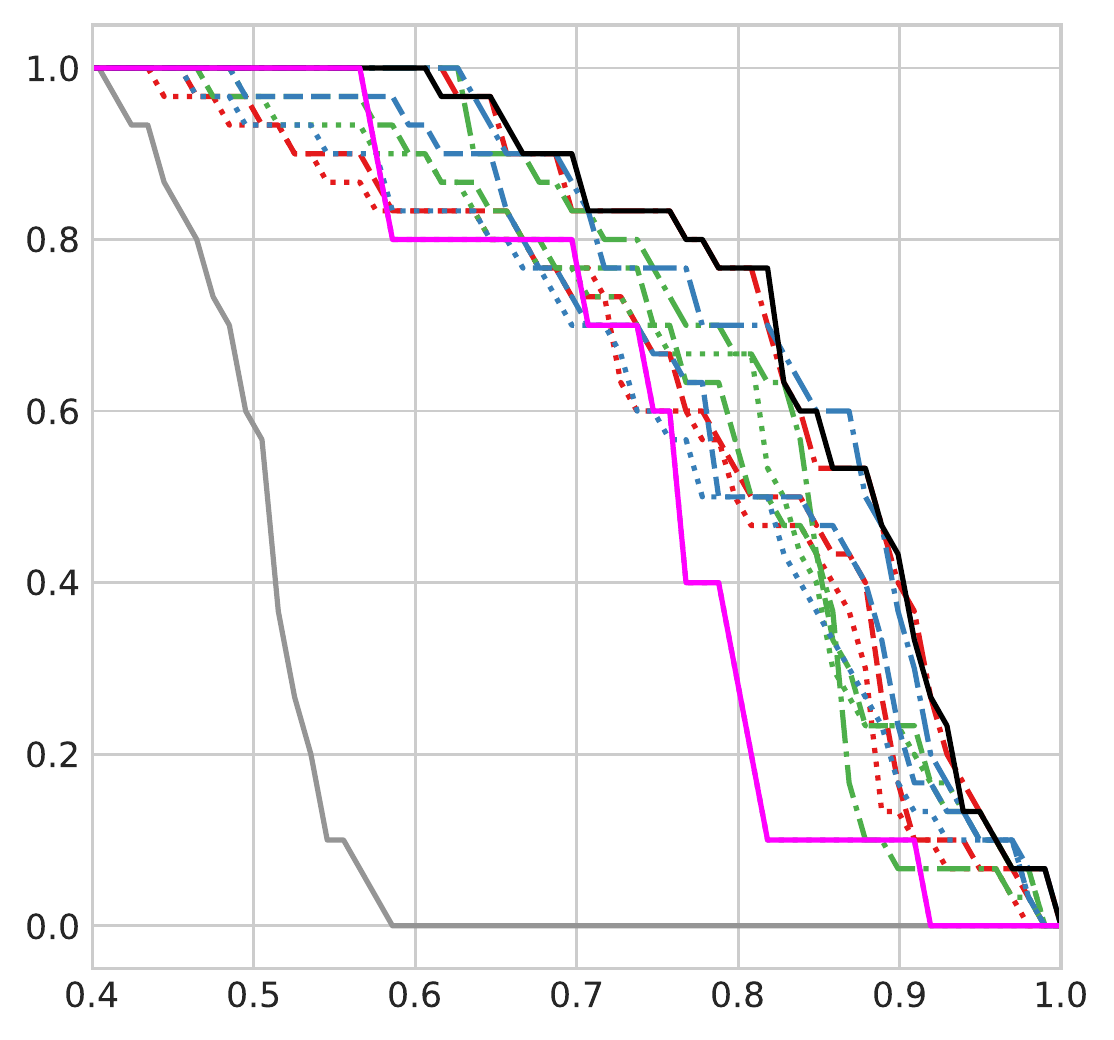}};
          \node (B) at ($(A.south)!-.03!(A.north)$) {\tiny Threshold };
          \node[rotate=90] (C) at ($(A.west)!-.03!(A.east)$) {\tiny Share};
        \end{tikzpicture}
        
        \caption{\shortstack{Group 3\\\musicgenre{}}}
    \end{subfigure}
    \hfill
    \begin{subfigure}[t]{0.22\textwidth}
        \centering
        \begin{tikzpicture}
          \node[inner sep=0pt] (A) {\includegraphics[width=\textwidth]{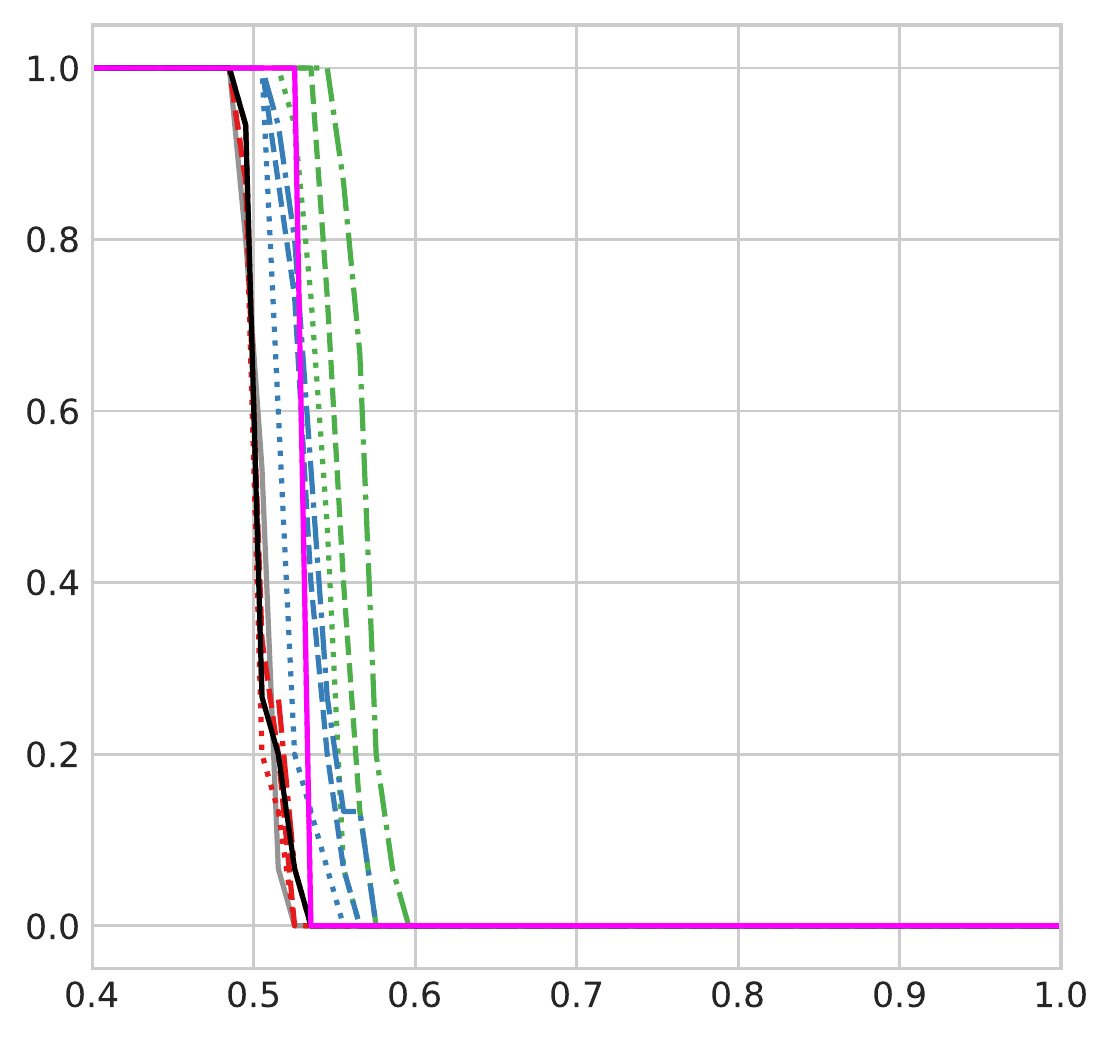}};
          \node (B) at ($(A.south)!-.03!(A.north)$) {\tiny Threshold };
          \node[rotate=90] (C) at ($(A.west)!-.03!(A.east)$) {\tiny Share};
        \end{tikzpicture}
        
        \caption{\shortstack{Group 3\\\sentimentpolarity{}}}
    \end{subfigure}
    \caption{\ROCAUC{} profiles for real datasets from groups 2 and 3. Colors represent the same legend as in figure \ref{fig:artifitial_profiles}.}
    \label{fig:real_profiles}
\end{figure*}

Supplementary \ROCAUC{} profiles are presented in figures \ref{fig:artifitial_profiles} and \ref{fig:real_profiles}. Overall, \mfgpc{} has a good performance, except \sentimentpolarity{} dataset. Notably, on this dataset all GPs-based methods have poor performance, which is not surprising, since we used a translation-invariant isotropic kernel, which is not suited well for highly clustered non-stationary data.

\input{data/fig_artificial_ROCAUC.tex}
\input{data/fig_real_ROCAUC.tex}

 \begin{table*}[t]
    \centering
    \caption{Average \ROCAUC{} among multiple runs on datasets from group 3 with natural noise.}
    \label{tab:results_group_3}
    \begin{adjustbox}{width=1\textwidth}
    \begin{tabular}{l|ccccccccccc}
    \hline
    & \mfgpc{} & \gpc{} & \logit{} & \xgb{}& \concMF{} \gpc{} & \concMF{} \logit{} & \concMF{} \xgb{} & \stackedMF{} \gpc{} & \stackedMF{} \logit{} & \stackedMF{} \xgb{} & \gpma{} \\
    \hline
    \musicgenre{} & $\mathbf{0.851}$ & 0.772 & 0.794 & 0.773 & $\mathbf{0.849}$ & 0.812 & $\mathbf{0.843}$ & 0.785 & 0.797 & 0.800 & 0.744 \\
    \sentimentpolarity{} & 0.504 & 0.502 & 0.542 & 0.520 & 0.505 & $\mathbf{0.569}$ & 0.538 & 0.504 & 0.553 & 0.533 & 0.531 \\
    \hline
    \end{tabular}
    \end{adjustbox}
    
\end{table*}


\subsection{Budget distribution among variable fidelity sources}

We studied how the ratio of low- and high-fidelity samples sizes affects the classification quality of \mfgpc{} on datasets from group 1. An experimental setup was the following: we assumed each high-fidelity entry cost X units, whereas low-fidelity entries cost a fraction of X (with various fractions for different experiments). Training samples were formed based on the total budget: some part of it was allocated for high-fidelity data, the rest was for low-fidelity data. If the whole budget was spent on high-fidelity data, then the training sample contained 100 entries. 

\begin{figure*}[t]
\centering
    {\small
          \begin{tabular}{|llll|}
            \hline
            LF cost: 
            \plainline{black} X/8 &
            \dashedline{black} X/4 &
            \dottedline{black} X/2 &
            \plainline{red} \gpc{} with only high-fidelity data \\
            \hline
            \end{tabular}
    
        }
    \begin{subfigure}[t]{0.22\textwidth}
        \centering
        \begin{tikzpicture}
          \node[inner sep=0pt] (A) {\includegraphics[width=\textwidth]{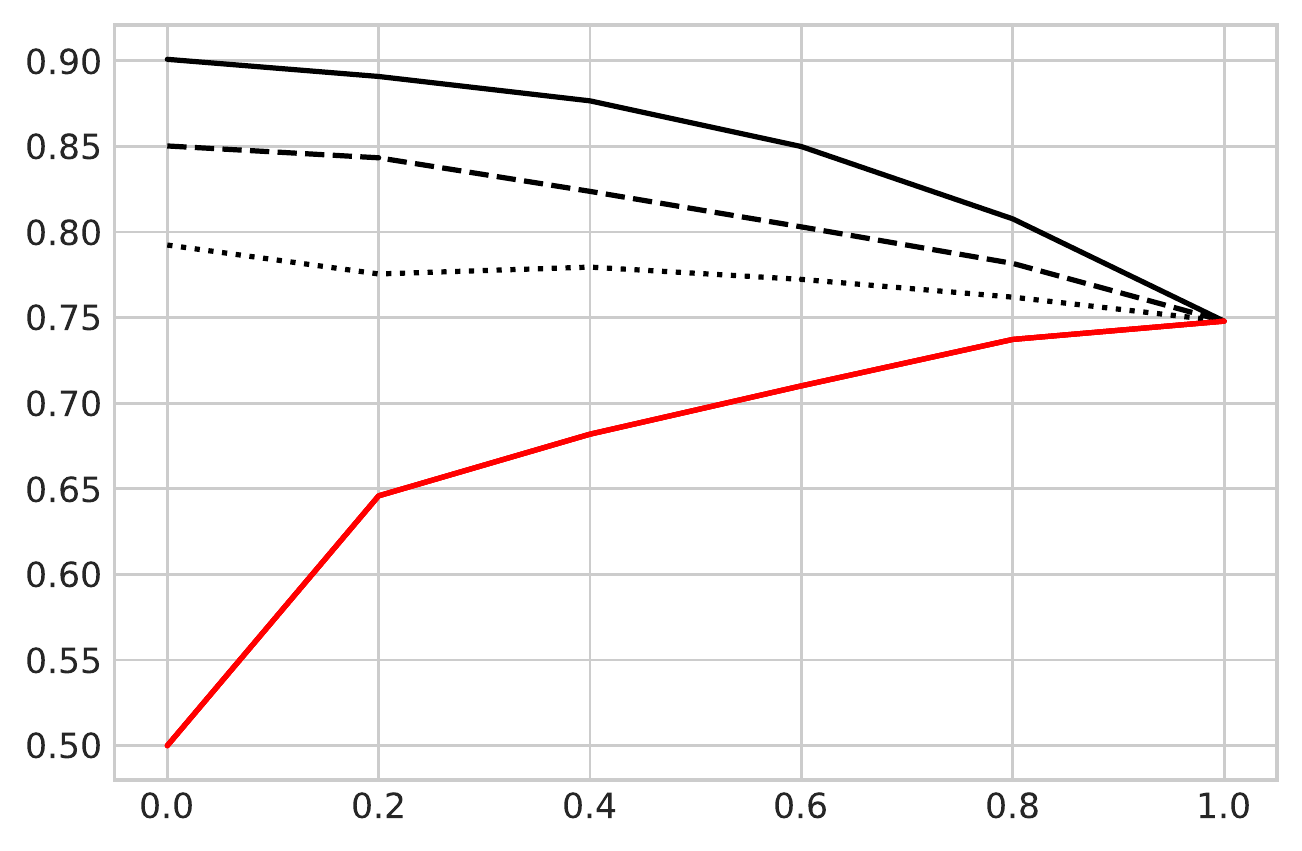}};
          \node (B) at ($(A.south)!-.03!(A.north)$) {\tiny HF share };
          \node[rotate=90] (C) at ($(A.west)!-.03!(A.east)$) {\tiny \ROCAUC{}};
        \end{tikzpicture}
        \caption{Noise level 0.0}
    \end{subfigure}%
    \hfill
    \begin{subfigure}[t]{0.22\textwidth}
        \centering
        \begin{tikzpicture}
          \node[inner sep=0pt] (A) {\includegraphics[width=\textwidth]{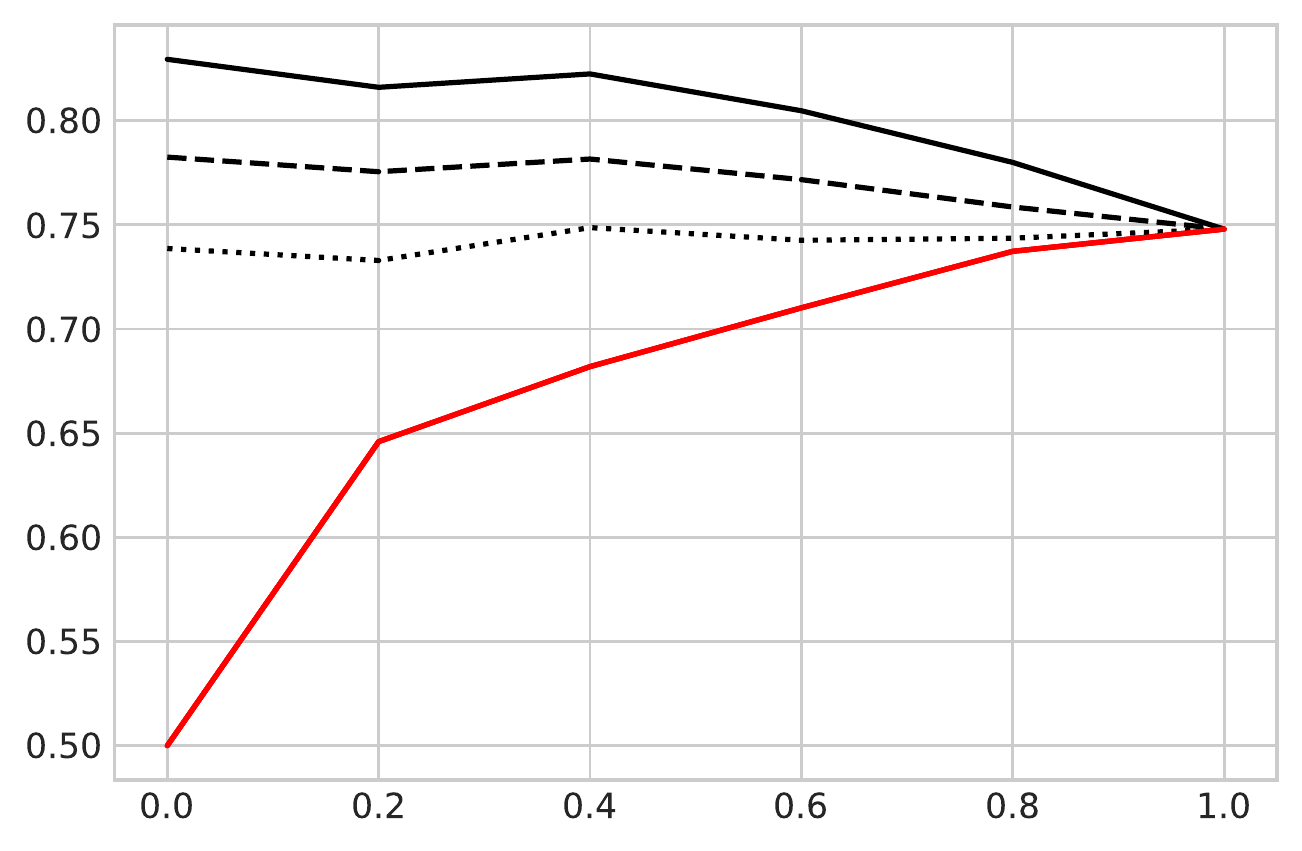}};
          \node (B) at ($(A.south)!-.03!(A.north)$) {\tiny HF share };
          \node[rotate=90] (C) at ($(A.west)!-.03!(A.east)$) {\tiny \ROCAUC{}};
        \end{tikzpicture}
        \caption{Noise level 0.2}
    \end{subfigure}%
    \hfill
    \begin{subfigure}[t]{0.22\textwidth}
        \centering
        \begin{tikzpicture}
          \node[inner sep=0pt] (A) {\includegraphics[width=\textwidth]{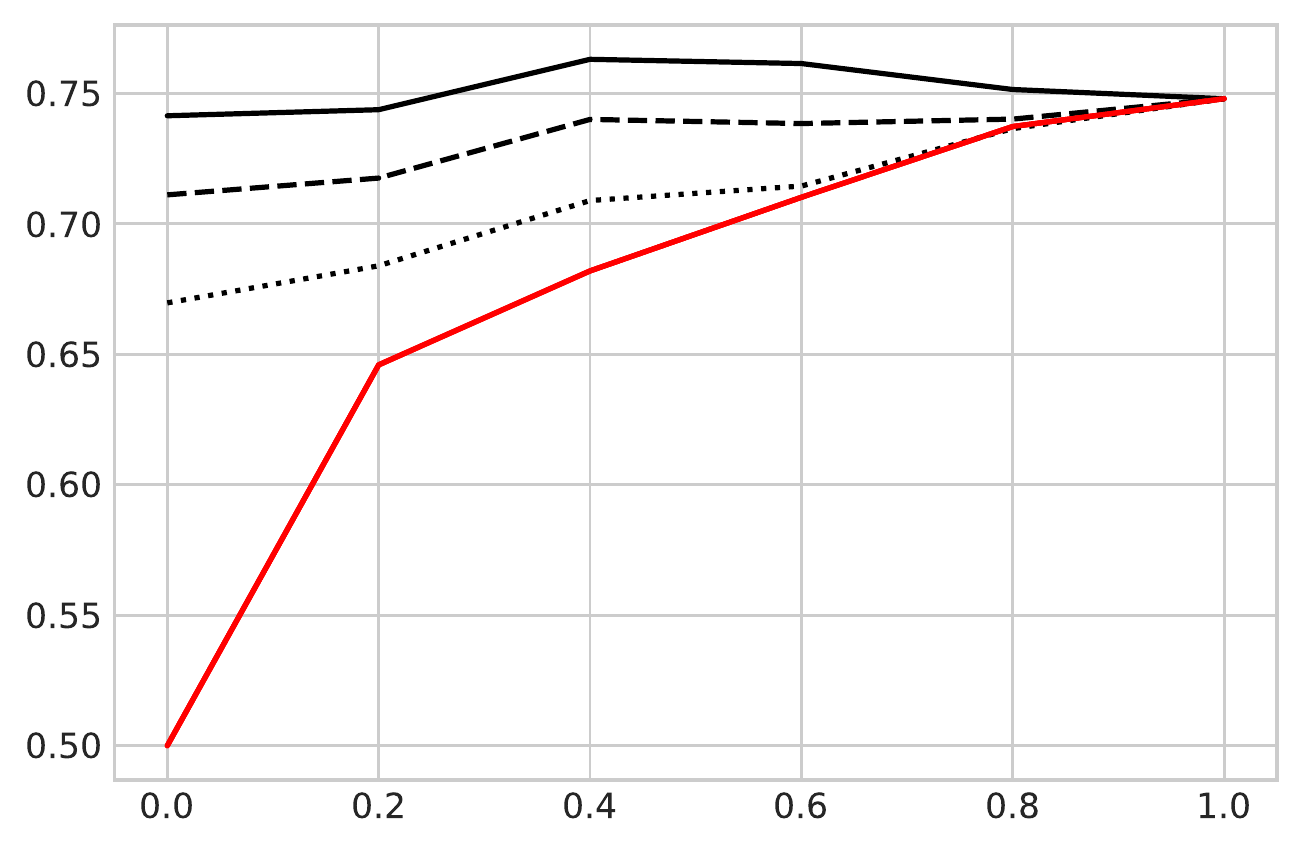}};
          \node (B) at ($(A.south)!-.03!(A.north)$) {\tiny HF share };
          \node[rotate=90] (C) at ($(A.west)!-.03!(A.east)$) {\tiny \ROCAUC{}};
        \end{tikzpicture}
        \caption{Noise level 0.3}
    \end{subfigure}%
    \hfill
    \begin{subfigure}[t]{0.22\textwidth}
        \centering
        \begin{tikzpicture}
          \node[inner sep=0pt] (A) {\includegraphics[width=\textwidth]{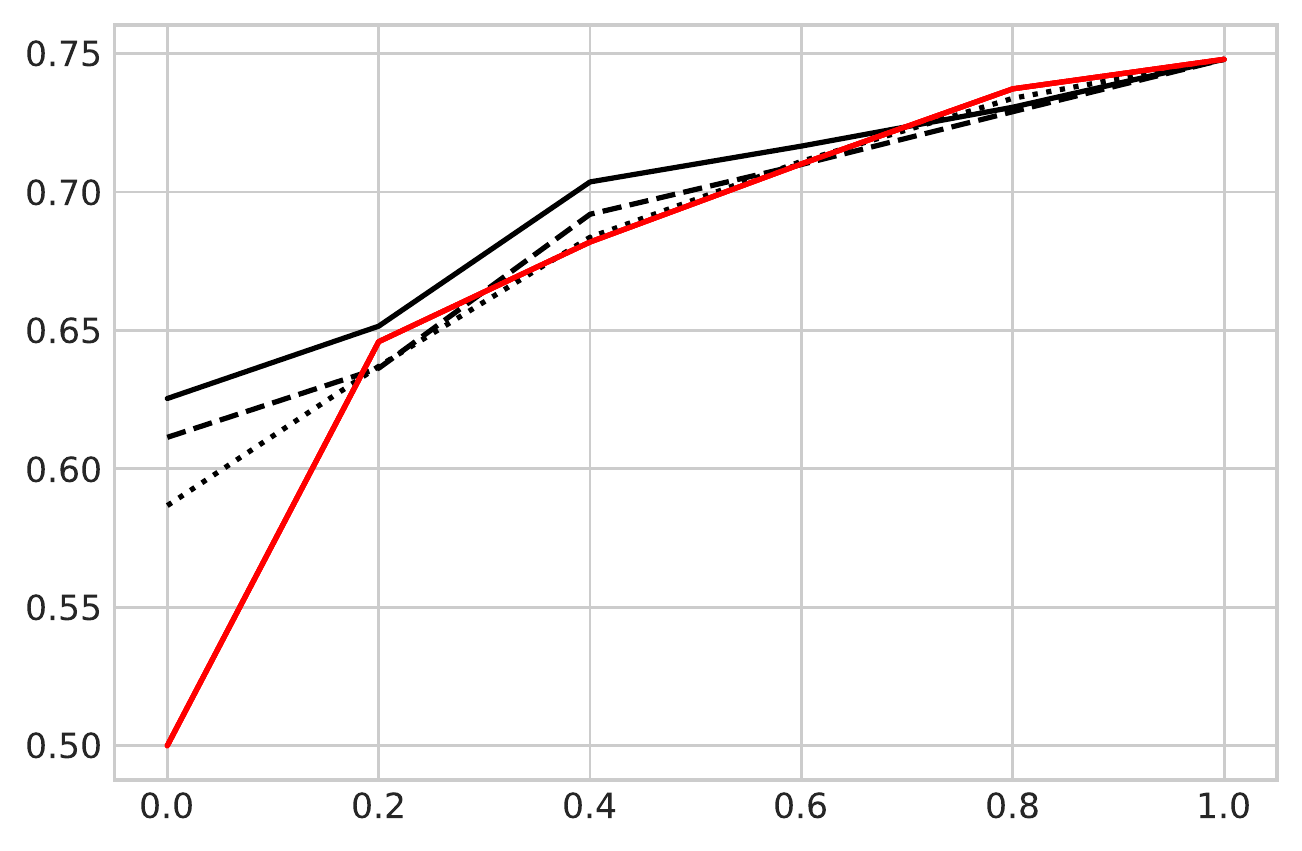}};
          \node (B) at ($(A.south)!-.03!(A.north)$) {\tiny HF share };
          \node[rotate=90] (C) at ($(A.west)!-.03!(A.east)$) {\tiny \ROCAUC{}};
        \end{tikzpicture}
        \caption{Noise level 0.4}
    \end{subfigure}%
    \hfill
    \caption{Performance of \mfgpc{} depending on share of budget allocated to high-fidelity data (HF share) for different ratios of low-fidelity cost to high-fidelity cost.}
    \label{fig:budget_lf_cost}
\end{figure*}

Figure \ref{fig:budget_lf_cost} demonstrates the more low-fidelity data is available or the less noise is in it, the better classifier works w.r.t. fixed amount of high-fidelity entries. That is, having fixed $D_H$, adding more data to $D_L$ with the same noise level in low-fidelity labels does not reduce the quality of predictions of our classifier. In the worst-case scenario, when low-fidelity labels are independent of high-fidelity ones, for example they consist of merely random noise, \mfgpc{} model degenerates to an ordinary \gpc{} trained on $D_H$, because in co-kriging formula \eqref{eq:cokriging_dependency} component $\rho f_L(x^H_i)$ becomes $0$, thus $f_H(x^H_i) = \delta(x^H_i)$. 

\begin{figure*}[t]
\centering
    {\small
          \begin{tabular}{|llll|}
            \hline
            Noise level: 
            \plainline{budget_noise_0.0} 0.0 &
            \plainline{budget_noise_0.2} 0.2 &
            \plainline{budget_noise_0.3} 0.3 &
            \plainline{budget_noise_0.4} 0.4 \\
            \hline
            \end{tabular}
    
    }
    \begin{subfigure}[t]{0.31\textwidth}
        \centering
        \begin{tikzpicture}
          \node[inner sep=0pt] (A) {\includegraphics[width=\textwidth]{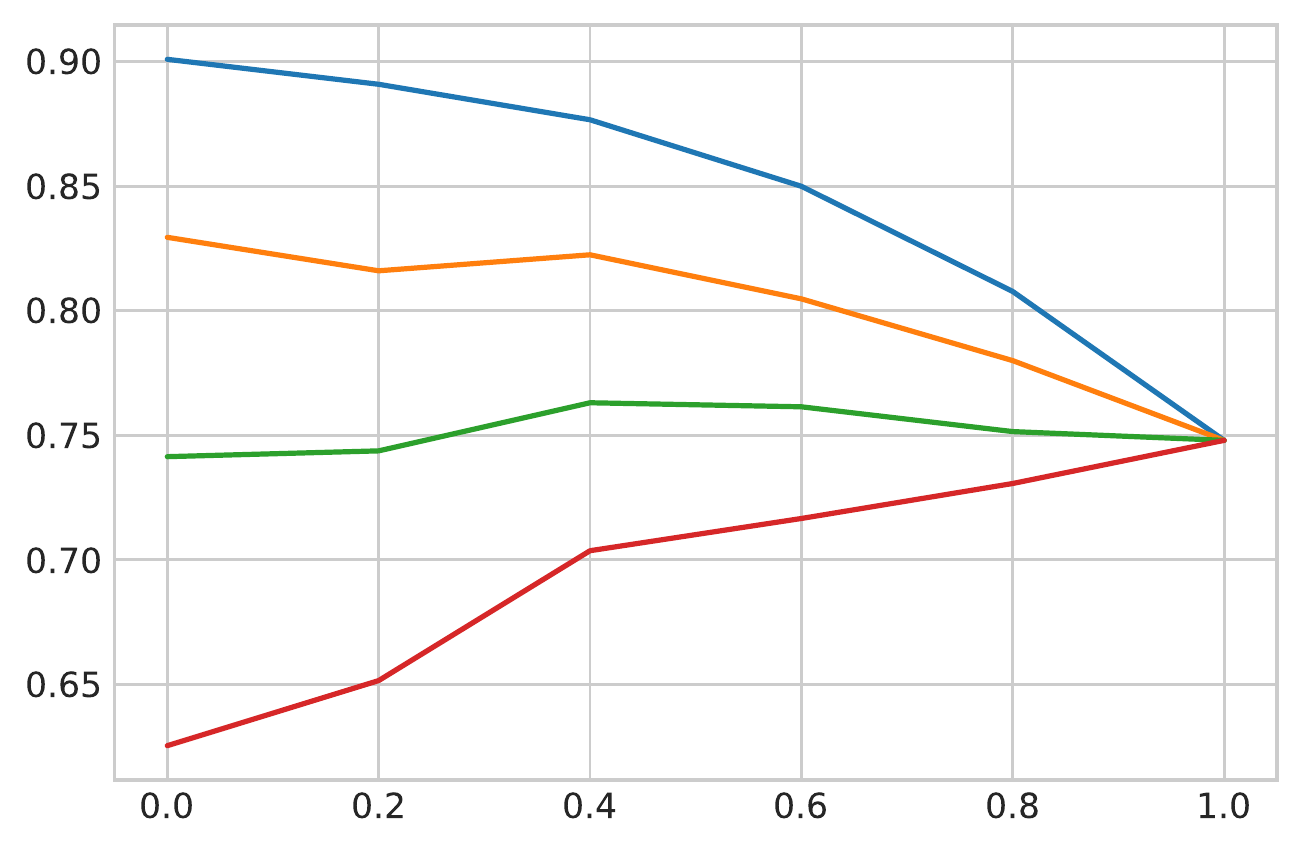}};
          \node (B) at ($(A.south)!-.03!(A.north)$) {\tiny HF share };
          \node[rotate=90] (C) at ($(A.west)!-.03!(A.east)$) {\tiny \ROCAUC{}};
        \end{tikzpicture}
        \caption{LF cost X/8}
    \end{subfigure}%
    \hfill
    \begin{subfigure}[t]{0.31\textwidth}
        \centering
        \begin{tikzpicture}
          \node[inner sep=0pt] (A) {\includegraphics[width=\textwidth]{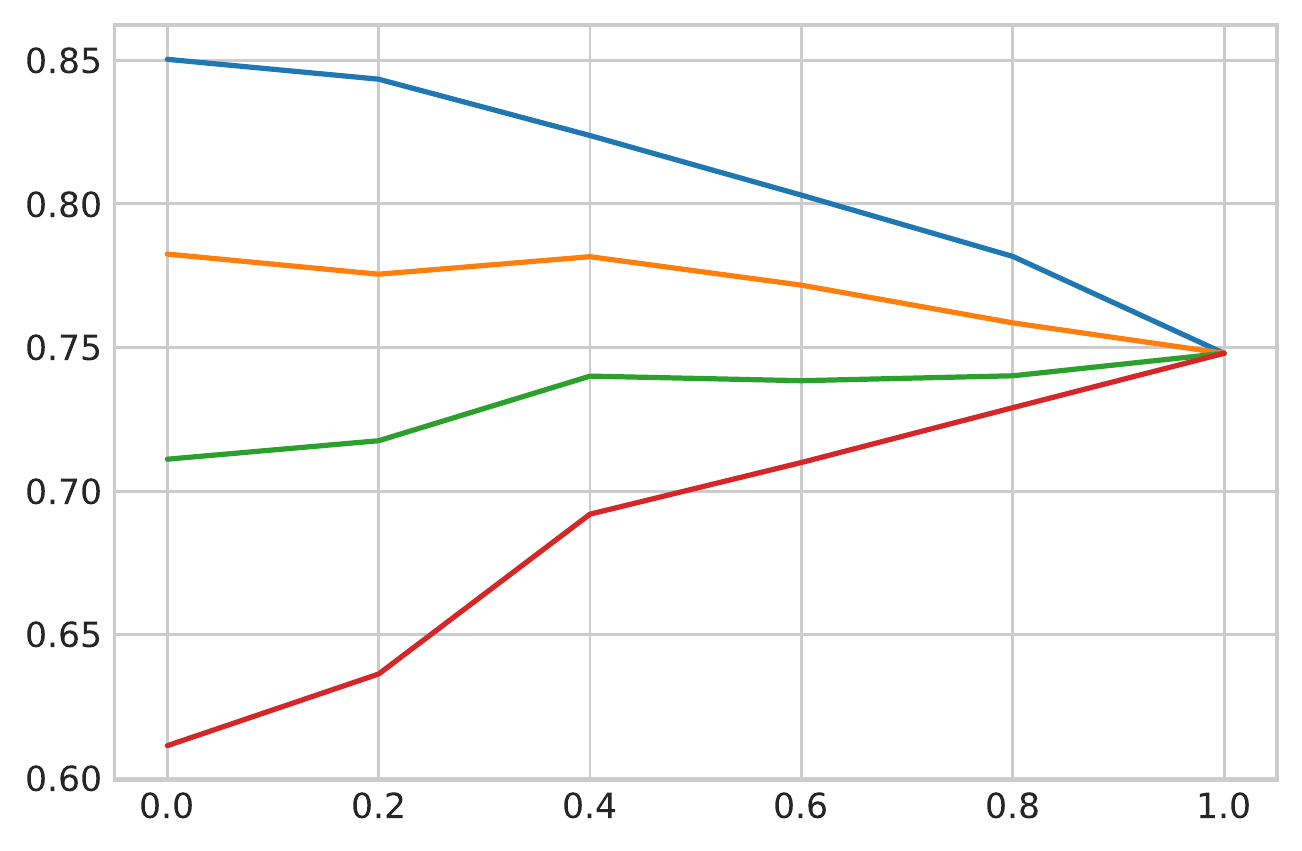}};
          \node (B) at ($(A.south)!-.03!(A.north)$) {\tiny HF share };
          \node[rotate=90] (C) at ($(A.west)!-.03!(A.east)$) {\tiny \ROCAUC{}};
        \end{tikzpicture}
        \caption{LF cost X/4}
    \end{subfigure}%
    \hfill
    \begin{subfigure}[t]{0.31\textwidth}
        \centering
        \begin{tikzpicture}
          \node[inner sep=0pt] (A) {\includegraphics[width=\textwidth]{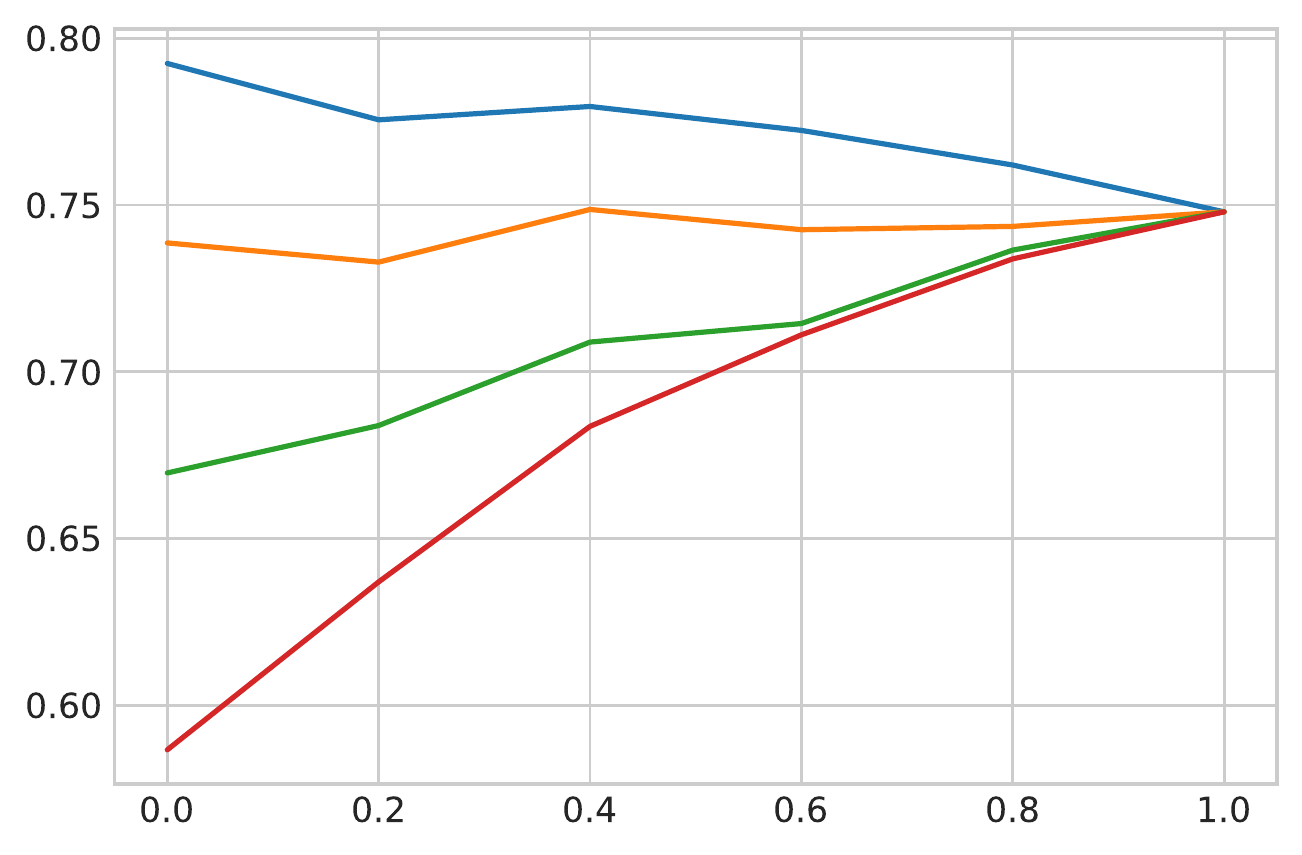}};
          \node (B) at ($(A.south)!-.03!(A.north)$) {\tiny HF share };
          \node[rotate=90] (C) at ($(A.west)!-.03!(A.east)$) {\tiny \ROCAUC{}};
        \end{tikzpicture}
        \caption{LF cost X/2}
    \end{subfigure}%
    \hfill
    \caption{Performance of \mfgpc{} depending on share of budget allocated to high-fidelity data (HF share) for different noise levels in low-fidelity labels.}
    \label{fig:budget_noise}
\end{figure*}

Figure \ref{fig:budget_noise} shows that in case of low noise level in low-fidelity labels sample size advantage overbalances decreased labels quality, thus spending all budget on low-fidelity data is the best option for this case. On the other hand, when the noise in low-fidelity is high, adding any amount of low-fidelity entries instead of high-fidelity ones to training sample reduces the performance of the classifier. 

These experiments show that in boundary cases single-fidelity \gpc{} is the choice either for training on low-fidelity data when the noise in labels is low or for training on high-fidelity data when noise is high, whereas \mfgpc{} works slightly better for intermediate noise levels in low-fidelity. It is not trivial to find the right balance in advance, but observations in this section can be used as a rule of thumb in practice.

\subsection{Sensitivity to hyperparameters}
We used radial basis functions as kernels for Gaussian processes in the following form:
\begin{equation}
\label{eq:rbf_kernel}
k_*(x_i, x_j) = exp(s_*)\ exp\left(-\frac{1}{2} \frac{\|x_i - x_j\|^2}{\sigma_*^2}\right),
\end{equation}
where $(s_*, \sigma_*)=\theta_*$ are kernel parameters, $* \in \{l, d\}$ indicates the corresponding Gaussian process. 

In these series of experiments we first tuned model on training samples, then varied some hyperparameters while kept others fixed to their values obtained during the training. While $\rho$ was varied, parameters of kernels were fixed. While $\theta_l=(s_l, \sigma_l)$ was varied across the grid of $s_l$ and $\sigma_l$ values, parameters of $k_d$ and $\rho$ were fixed and vice versa for $\theta_d=(s_d, \sigma_d)$. Eventually, for each combination of hyperparameters we estimated model's performance on the corresponding validation samples.

\begin{figure*}[t]
    \centering
    \begin{subfigure}[t]{0.28\textwidth}
        \centering
        \begin{tikzpicture}
          \node[inner sep=0pt] (A) {\includegraphics[width=\textwidth]{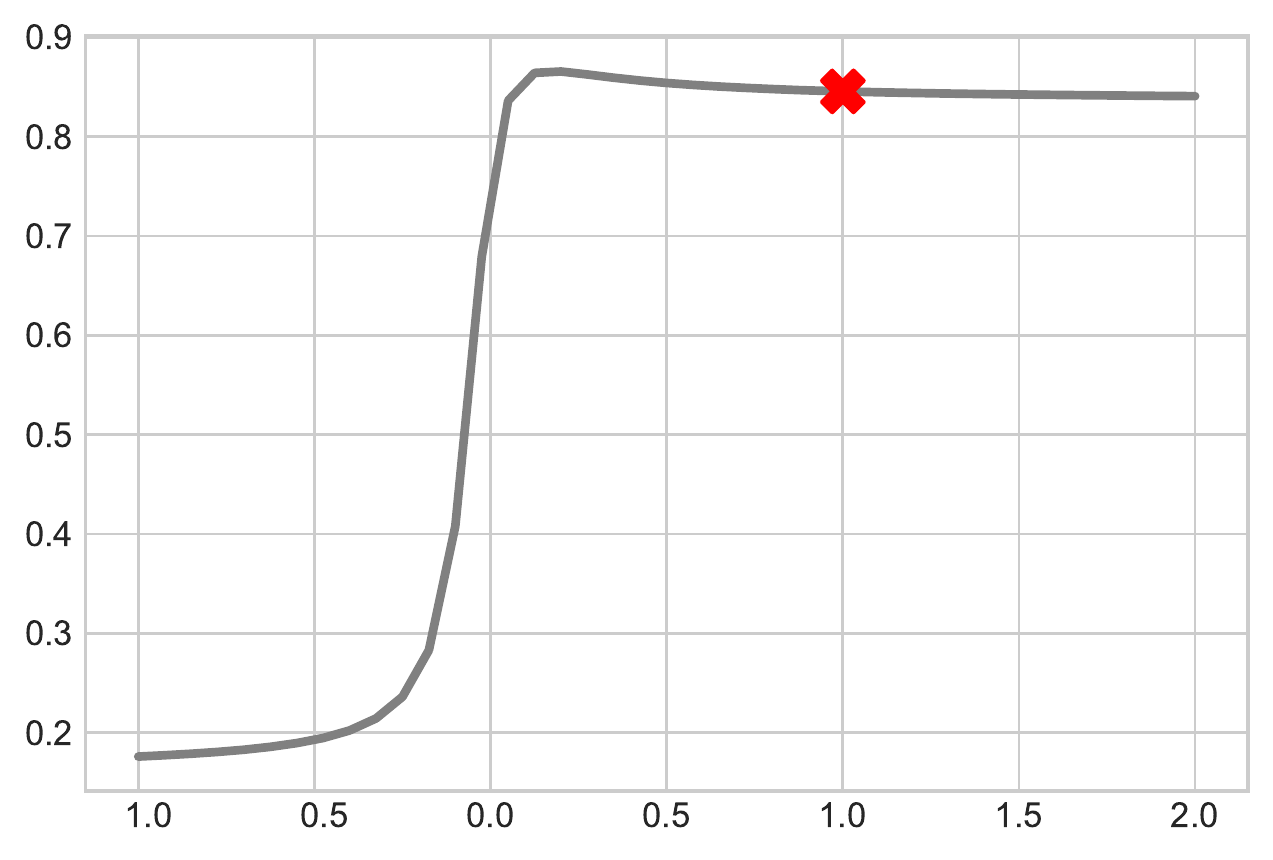}};
          \node (B) at ($(A.south)!-.03!(A.north)$) {\small $\rho$ };
          \node[rotate=90] (C) at ($(A.west)!-.03!(A.east)$) {\tiny \ROCAUC{}};
        \end{tikzpicture}
        
        \caption{\shortstack{linear coefficient\\in co-kriging}}
    \end{subfigure}
    \hfill
    \begin{subfigure}[t]{0.28\textwidth}
        \centering
        \begin{tikzpicture}
          \node[inner sep=0pt] (A) {\includegraphics[trim={0 0 4cm 0},clip, width=\textwidth]{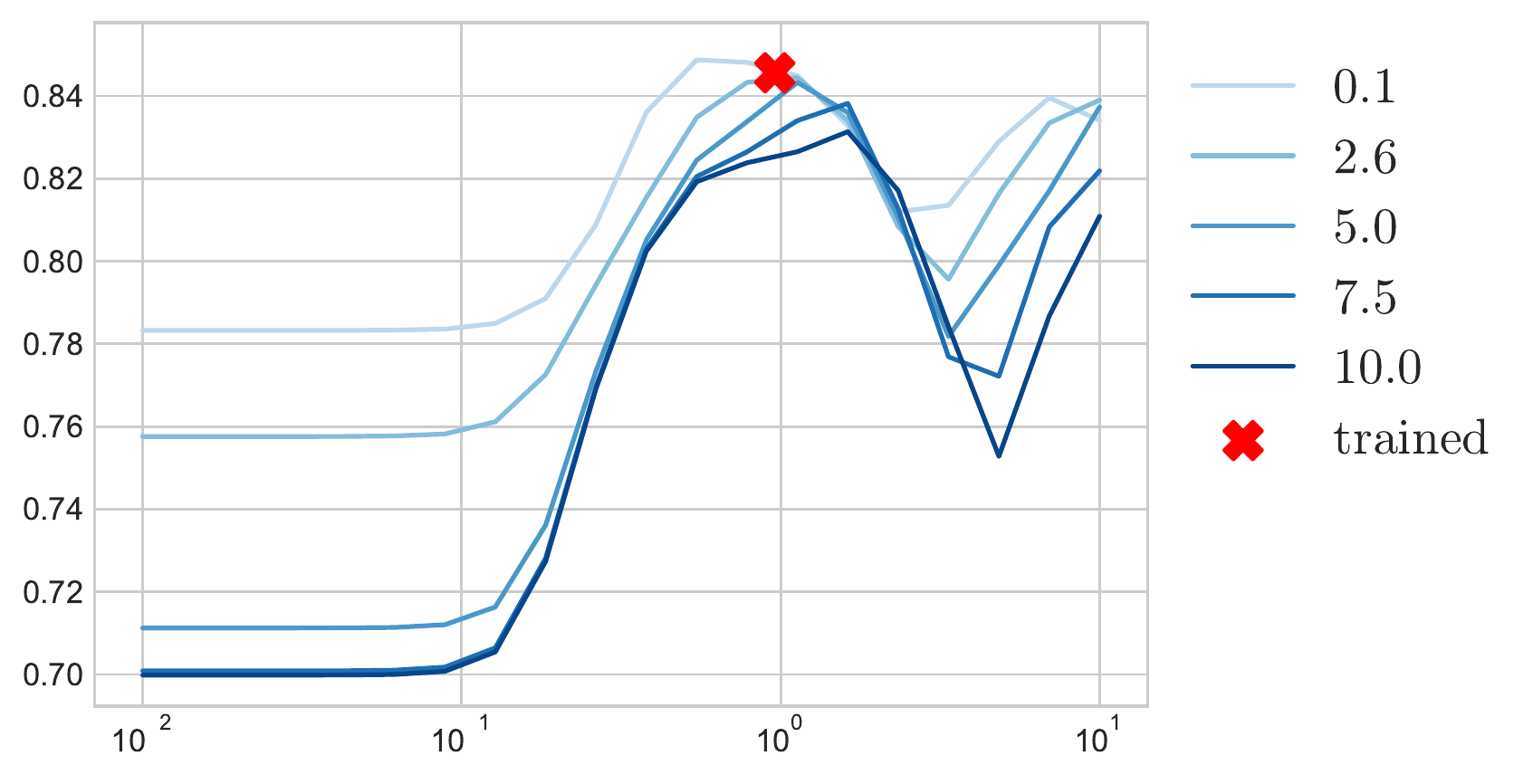}};
          \node (B) at ($(A.south)!-.03!(A.north)$) {\small $\sigma_l$ };
          \node[rotate=90] (C) at ($(A.west)!-.03!(A.east)$) {\tiny \ROCAUC{}};
        \end{tikzpicture}
        
        \caption{\shortstack{$k_l(\cdot, \cdot)$\\hyperparams}}
        \label{subfig:sensetivity_k_l}
    \end{subfigure}
    \hfill
    \begin{subfigure}[t]{0.4\textwidth}
        \centering
        \begin{tikzpicture}
          \node[inner sep=0pt] (A) {\includegraphics[width=\textwidth]{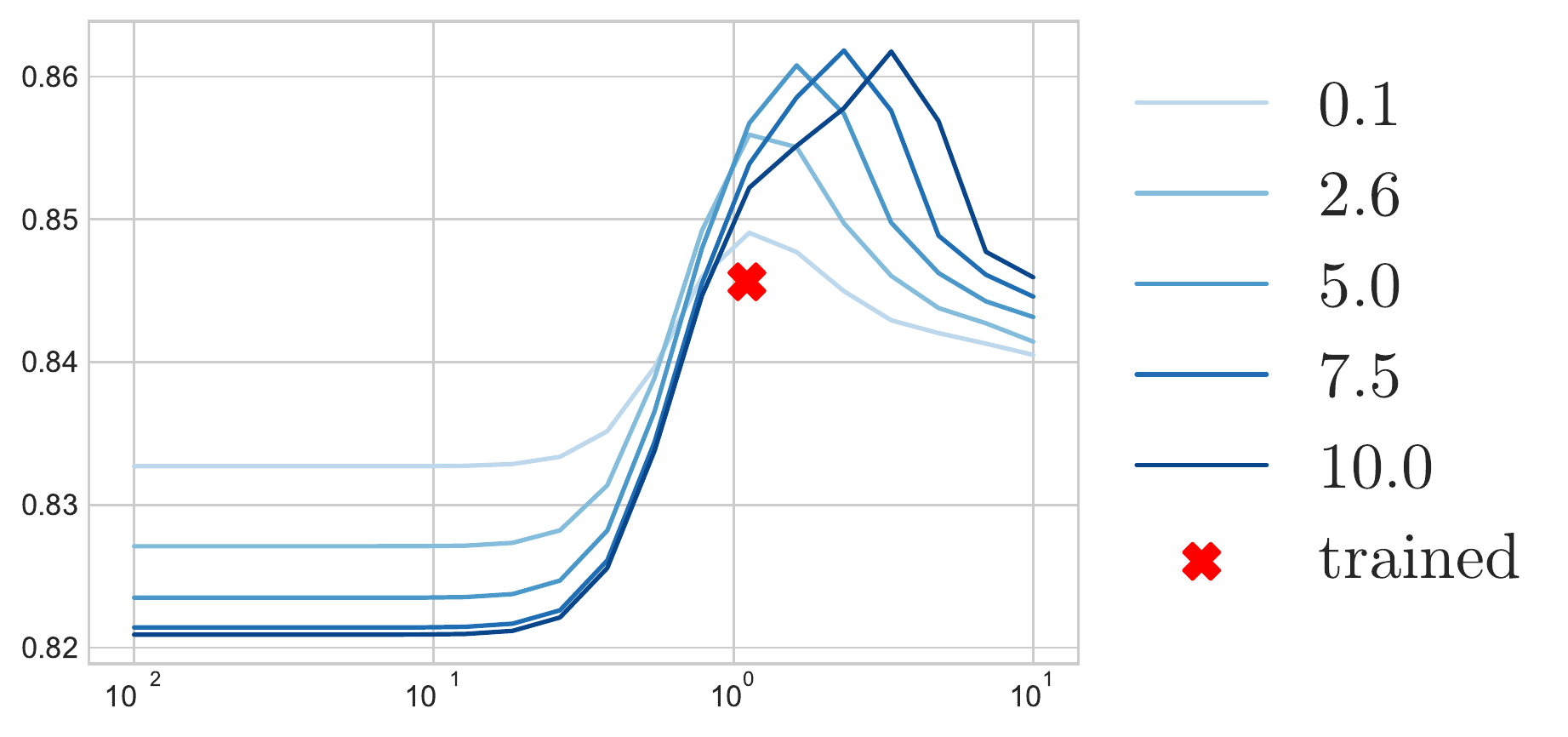}};
          \node (B) at ($(A.south)!-.03!(A.north) + (A.west)!+.38!(A.east)$) {\small $\sigma_d$ };
          \node[rotate=90] (C) at ($(A.west)!-.03!(A.east)$) {\tiny \ROCAUC{}};
        \end{tikzpicture}
        
        \caption{\shortstack{$k_d(\cdot, \cdot)$\\hyperparams}}
        \label{subfig:sensetivity_k_d}
    \end{subfigure}
    
    \caption{Sensitivity of model performance to its hyperparameters in case of low or moderate noise in low-fidelity labels. Curves of different shades in figures \ref{subfig:sensetivity_k_l} and \ref{subfig:sensetivity_k_d} are associated with the the log-scale coefficient ($s_*$ in \eqref{eq:rbf_kernel}) of the corresponding kernel. Red mark indicates parameters and performance of the tuned model during the training.}
    \label{fig:sensitivity}
\end{figure*}

\begin{figure*}[t]
    \centering
    \begin{subfigure}[t]{0.28\textwidth}
        \centering
        \begin{tikzpicture}
          \node[inner sep=0pt] (A) {\includegraphics[width=\textwidth]{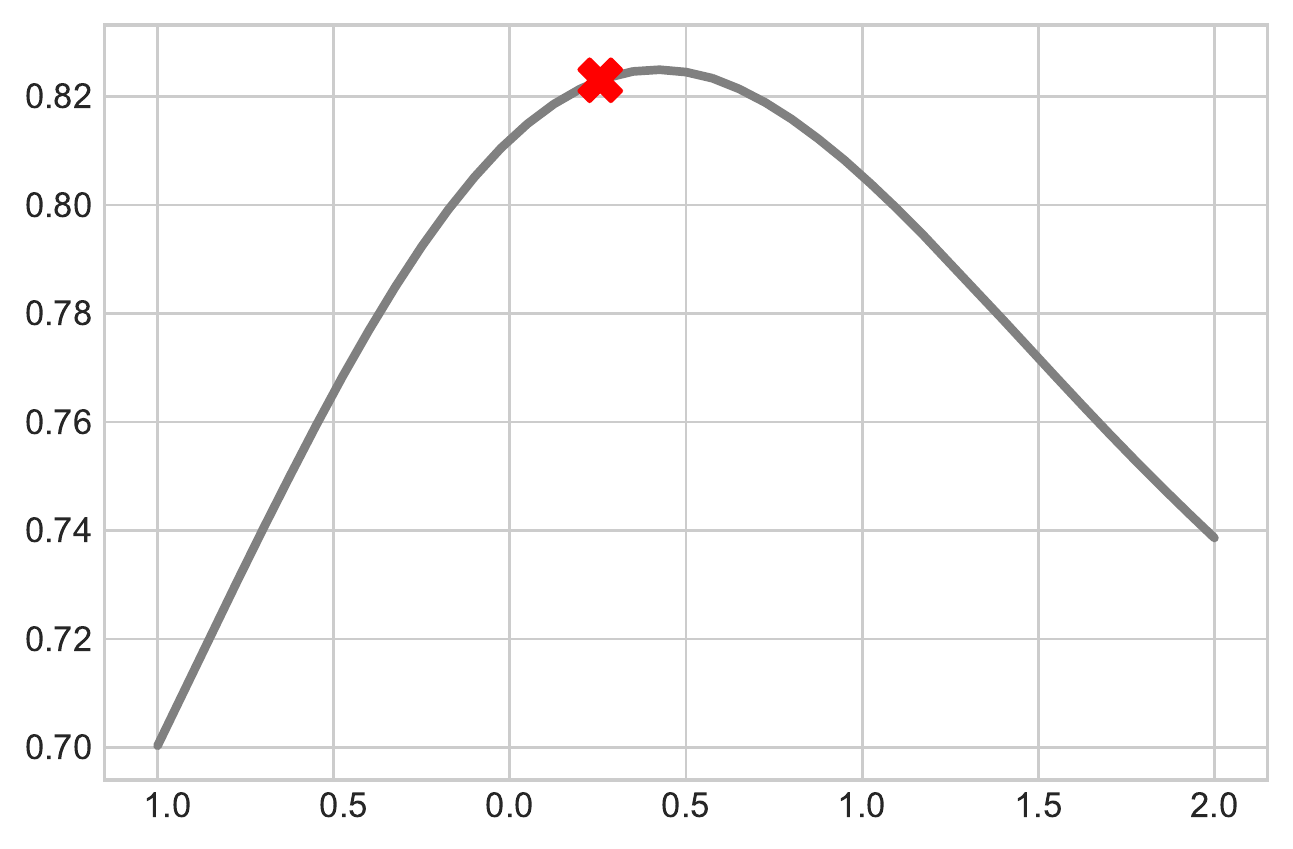}};
          \node (B) at ($(A.south)!-.03!(A.north)$) {\small $\rho$ };
          \node[rotate=90] (C) at ($(A.west)!-.03!(A.east)$) {\tiny \ROCAUC{}};
        \end{tikzpicture}
        
        \caption{\shortstack{linear coefficient\\in co-kriging}}
    \end{subfigure}
    \hfill
    \begin{subfigure}[t]{0.28\textwidth}
        \centering
        \begin{tikzpicture}
          \node[inner sep=0pt] (A) {\includegraphics[trim={0 0 5.5cm 0},clip, width=\textwidth]{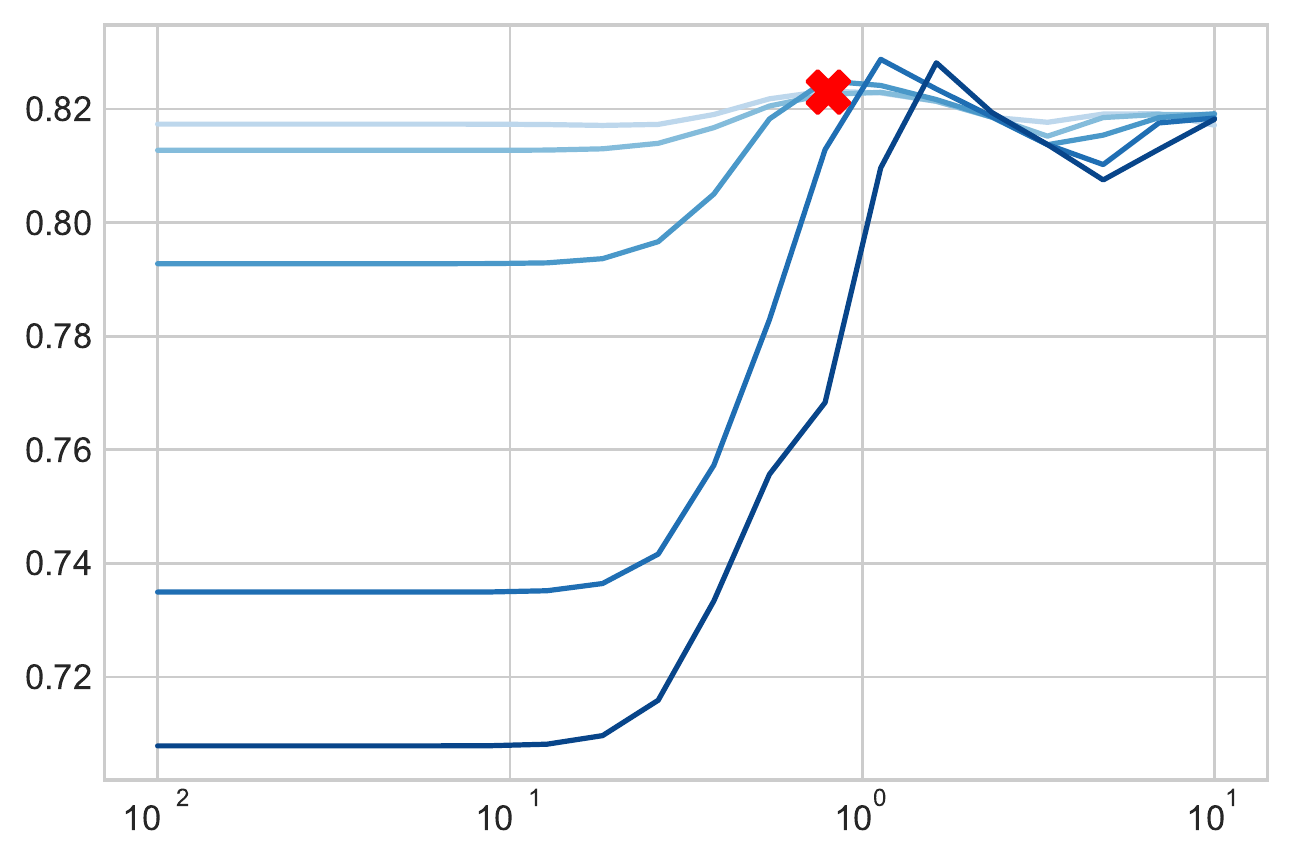}};
          \node (B) at ($(A.south)!-.03!(A.north)$) {\small $\sigma_l$ };
          \node[rotate=90] (C) at ($(A.west)!-.03!(A.east)$) {\tiny \ROCAUC{}};
        \end{tikzpicture}
        
        \caption{\shortstack{$k_l(\cdot, \cdot)$\\hyperparams}}
        \label{subfig:sensetivity_k_l_2}
    \end{subfigure}
    \hfill
    \begin{subfigure}[t]{0.4\textwidth}
        \centering
        \begin{tikzpicture}
          \node[inner sep=0pt] (A) {\includegraphics[width=\textwidth]{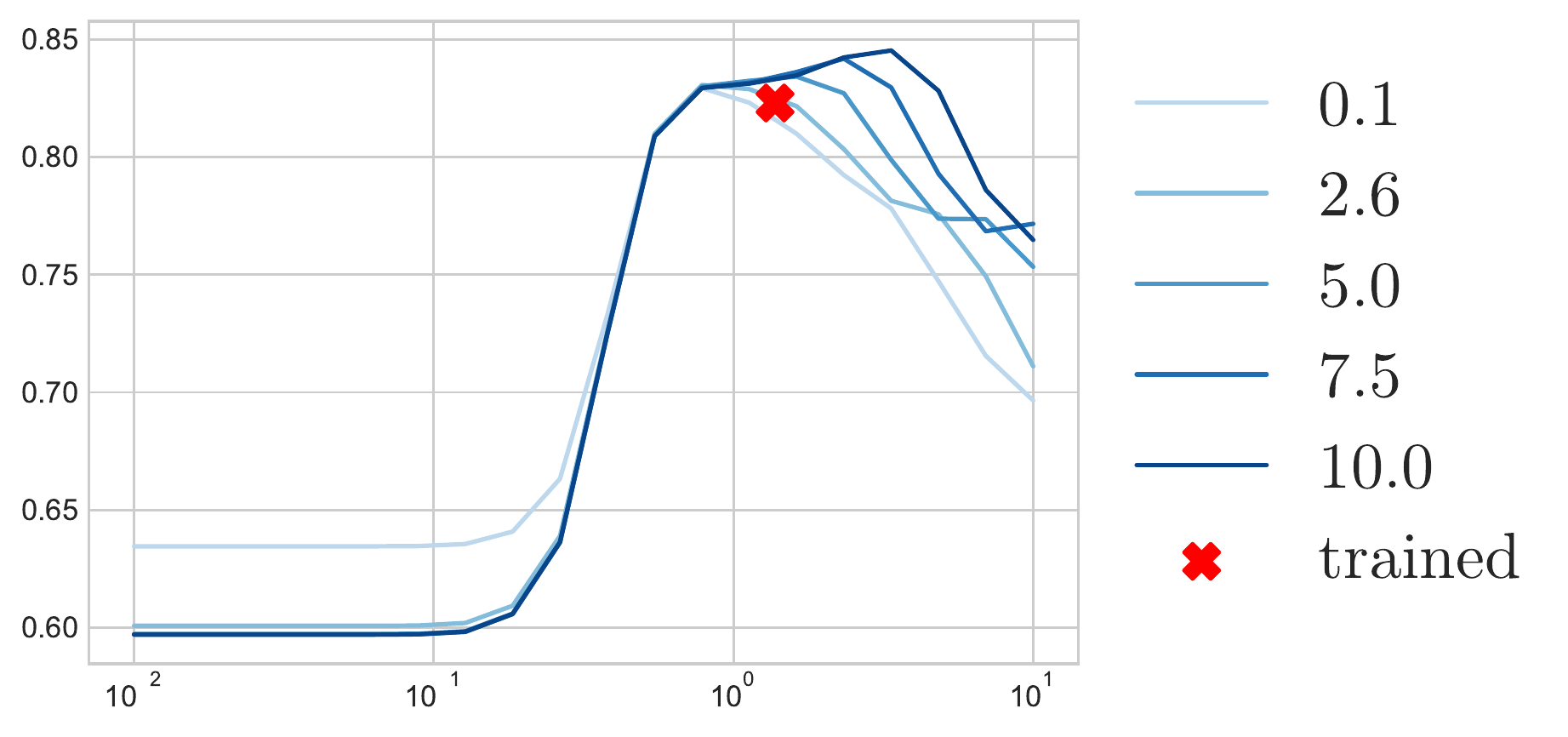}};
          \node (B) at ($(A.south)!-.03!(A.north) + (A.west)!+.38!(A.east)$) {\small $\sigma_d$ };
          \node[rotate=90] (C) at ($(A.west)!-.03!(A.east)$) {\tiny \ROCAUC{}};
        \end{tikzpicture}
        
        \caption{\shortstack{$k_d(\cdot, \cdot)$\\hyperparams}}
        \label{subfig:sensetivity_k_d_2}
    \end{subfigure}
    
    \caption{Sensitivity of model performance to its hyperparameters in case of high noise in low-fidelity labels. Curves of different shades in figures \ref{subfig:sensetivity_k_l_2} and \ref{subfig:sensetivity_k_d_2} are associated with the the log-scale coefficient ($s_*$ in \eqref{eq:rbf_kernel}) of the corresponding kernel. Red mark indicates parameters and performance of the tuned model during the training.}
    \label{fig:sensitivity_2}
\end{figure*}

Figures \ref{fig:sensitivity} and \ref{fig:sensitivity_2} show a typical sensitivity of model's performance on the validation set with respect to the hyperparameters $\rho, \theta_l, \theta_d$ for cases with low or moderate noise  and case with high noise in low-fidelity data respectively. The former cases are characterized with low local sensitivity to $\rho$ and a sharp decrease in performance when its sign changes; performance is also more sensitive to parameters of $k_l$ than to those of $k_d$. For latter cases the situation is opposite: the performance is more affected by local changes in $\rho$, regarding kernels the model is vice versa more sensitive to parameters of $k_d$ than those of $k_l$.

\section{Discussions}
\label{sec:discussions}

Despite the algorithm was proposed for two levels of fidelities, it can be trivially generalized to an arbitrary number of levels assuming the Markov property of fidelity-levels \cite{kennedy2000predicting}.

Further research should be dedicated to the theoretical investigation of budget distribution among fidelities for optimizing the performance of multi-fidelity classifier over single-fidelity ones. The prior work \cite{pmlr-v54-zaytsev17a} has been successfully studied this issue for the regression problem, which resulted in the analytic formula for the optimal budget balance, although the case for classification problem looks more challenging.

Finally, in order to make our method applicable to a wide range of real projects, its scalability should be improved. A number of approaches to do this has been recently reviewed \cite{Zaytsev2017,liu2018gaussian}.

\section{Conclusions}
\label{sec:conclusions}

Multi-fidelity modeling of discrete response surfaces can be put to good use in a number of applied disciplines, yet such methods have got little attention so far. In this work, we extended Laplace inference algorithm for classification based on GPs to make it work with multi-fidelity data. By modeling latent GPs dependency with a co-kriging schema, which has been used previously for multi-fidelity regression, our method can identify not only the overall relevance of low-fidelity data, but resolve local item-dependent discrepancies between fidelities due to inference on residual Gaussian process $\delta$.  

We evaluated our method on multiple artificial and real datasets with natural and various levels of simulated noise and compared its performance with a number of baseline approaches and state-of-the-art methods. 
We also experimentally studied under which conditions adding noisy low-fidelity to the training set increases quality on top of high-fidelity data classification.
Depending on the dataset nature, \mfgpc{} can alternate its performance with respect to other methods, however, it is more resistant to different noise levels in low-fidelity labels. That is, when the classifiers based on GPs can learn datasets well, \mfgpc{} has a top performance, whereas in other cases our method is on par with the considered methods. 


\clearpage

\appendix
\clearpage
\section{Theoretical and experimental details}
\label{sec:appendix}

This appendix contains detailed information about some key identities and experimental setup. We also published source code for our model and experiments in this repository \texttt{https://github.com/user525/mfgpc}.

\begin{table}[ht]
    \centering
    \begin{adjustbox}{width=1\textwidth}
    \begin{tabular}{|c|l|c|}
    \hline
    Notations & Descriptions & Specification \\
    \hline
        $\Omega$ & the measurable domain of data & $\Omega \subset \mathbb{R}^d$ \\
        $c(\cdot)$ & a binary function defined on $\Omega$ & - \\
        $D_H$ & high-fidelity sample & $D_H = \{(x^H_i, y^H_i)\}_{i=1}^{n_H}$ \\
        $D_L$ & low-fidelity sample & $D_L = \{(x^L_i, y^L_i)\}_{i=1}^{n_L}$ \\
        $X_H$ & points of high-fidelity sample & $X_H = \{x^H_i\}_{i=1}^{n_H}$\\
        $X_L$ & points of low-fidelity sample & $X_L = \{x^L_i\}_{i=1}^{n_L}$\\
        $f_H(\cdot)$ & latent Gaussian Process for high-fidelity & $f_H(x) = \rho f_L(x) + \delta(x)$ \\
        $f_L(\cdot)$ & latent Gaussian Process for low-fidelity & - \\
        $\delta(\cdot)$ & latent residual Gaussian Process & - \\
        $\rho$ & linear coefficient for co-kriging dependency & $\rho \in \mathbb{R}$ \\ 
        $k_l(\cdot, \cdot)$ & prior kernel for $f_L(\cdot)$ & - \\
        $k_d(\cdot, \cdot)$ & prior kernel for $\delta(\cdot)$ & - \\
        $\theta_l$ & parameters of kernel $k_l$ & a multi-dimensional real vector \\
        $\theta_d$ & parameters of kernel $k_d$ & a multi-dimensional real vector \\
        $\sigma(\cdot)$ & sigmoid function & $\sigma(z) = \frac{1}{1 + exp(-z)}$ \\
        $\omega(\cdot)$ & first derivative of $\sigma(\cdot)$ & $\omega(z) = \sigma(z)(1 - \sigma(z))$ \\
        $\zeta(\cdot)$ & second derivative of $\sigma(\cdot)$ & $\zeta(x) = \sigma(x)(1 - \sigma(x))(1 - 2\sigma(x))$ \\
        $\lambda$ & log-likelihood & $\log p(\mathbf{y}^L, \mathbf{y}^H|\boldsymbol{\xi})$ \\
        $\mathcal{L}$ & approximate log marginal likelihood & $\log \widetilde{q}(\mathbf{y}^L, \mathbf{y}^H|X_L, X_H, \rho, \theta_l, \theta_d)$ \\
    \hline
    \end{tabular}
    \end{adjustbox}
    \vspace{3pt}
    \caption{Some of notations used in the paper.}
    \label{tab:notations}
\end{table}

\subsection{Correctness Of The Method}
\label{sec:correctness}

Optimization problem \eqref{eq:laplace_argmax} has a unique solution if $\Psi$ is concave. We prove it by showing that Hessian of $\Psi$ is negative semi-definite. The Hessian is 
\begin{equation}
\nabla\nabla\Psi(\boldsymbol{\xi}) = -W - K\inv,
\end{equation}
where K is positive semi-definite, since it is a kernel matrix. 

Matrices $A$ and $D$ in the definition of $W$ \eqref{eq:W} are positive semi-definite, because their diagonal elements are non-negative. The block of $W$ that contains $D$ can be represented via Kronecker product:
\begin{equation}
\label{eq:kronecker_product_D}
\begin{bmatrix}
\rho^2 D & \rho D \\
\rho D & D
\end{bmatrix} = \begin{bmatrix}
\rho^2 & \rho\\
\rho & 1
\end{bmatrix} \otimes D.
\end{equation}

Both multiplicands in \eqref{eq:kronecker_product_D} are positive semi-definite, thus, their Kronecker product is also positive semi-definite \cite{schacke2013kronecker}.

Hence, matrix $W$ is positive semi-definite, because it factorizes into two positive semi-definite blocks. Finally, the Hessian is negative semi-definite as a negation of sum of two positive semi-definite matrices.

\subsection{Inference For Equation \eqref{eq:d_log_det_B_d_rho}}
\label{seq:inference_d_log_det_B_d_rho}

\begin{equation}
\label{eq:d_log_det_B_d_rho_inference}
\begin{split}
&\frac{\partial \log |B|}{\partial \rho} = \textup{tr}\left(B\inv \frac{\partial B}{\partial \rho}\right) = \textup{tr}\left(B\inv \left(\frac{\partial W^{\frac{1}{2}}}{\partial \rho} K W^{\frac{1}{2}} + W^{\frac{1}{2}} K \frac{\partial W^{\frac{1}{2}}}{\partial \rho} \right)\right) = \\
&= \textup{tr}\left(B\inv \left(W^{-\frac{1}{2}} W^{\frac{1}{2}}\right) \frac{\partial W^{\frac{1}{2}}}{\partial \rho} K W^{\frac{1}{2}}\right) + \textup{tr}\left(B\inv W^{\frac{1}{2}} K \frac{\partial W^{\frac{1}{2}}}{\partial \rho} \left(W^{\frac{1}{2}} W^{-\frac{1}{2}}\right)\right) = \\
&= \textup{tr}\left(K W^{\frac{1}{2}} B\inv W^{-\frac{1}{2}} \left(W^{\frac{1}{2}} \frac{\partial W^{\frac{1}{2}}}{\partial \rho}\right) \right) + \textup{tr}\left(W^{-\frac{1}{2}}B\inv W^{\frac{1}{2}} K \left(\frac{\partial W^{\frac{1}{2}}}{\partial \rho} W^{\frac{1}{2}} \right) \right) = \\
&= \textup{tr}\left((K\inv + W)\inv  \frac{\partial  W}{\partial \rho}\right) = \sum_{\textup{all elements}} \left((K\inv + W)\inv \circ \frac{\partial  W}{\partial \rho}\right)
\end{split}
\end{equation}

The last line of \eqref{eq:d_log_det_B_d_rho_inference} is obtained because of the following identities:

\begin{equation}
\frac{\partial W^{\frac{1}{2}}}{\partial \rho} W^{\frac{1}{2}} + W^{\frac{1}{2}} \frac{\partial W^{\frac{1}{2}}}{\partial \rho} = \frac{\partial  W^{\frac{1}{2}}  W^{\frac{1}{2}}}{\partial \rho} = \frac{\partial  W}{\partial \rho}
\end{equation}

\begin{equation}
K W^{\frac{1}{2}} B\inv W^{-\frac{1}{2}} = K\left(W^{\frac{1}{2}} B\inv W^{\frac{1}{2}}\right) W\inv = K(K + W\inv)\inv W\inv = (K\inv + W)\inv
\end{equation}

\begin{equation}
W^{-\frac{1}{2}} W^{\frac{1}{2}} B\inv K  = W\inv \left(W^{\frac{1}{2}} B\inv W^{\frac{1}{2}}\right) K = W\inv(K + W\inv)\inv K = (K\inv + W)\inv
\end{equation}

\subsection{Components Of \eqref{eq:d_L_d_xi}}
\label{seq:components_of_d_L_d_xi}

Let us denote $M \overset{\mathsmaller{\triangle}}{=} (K\inv + W)\inv$.

For indices $i$ corresponding to low-fidelity data on $X_L$ ($i=1...n_l$):
\begin{equation}
\label{eq:MFGPC_partial_q_LL}
M_{i,i} \frac{\partial^3}{\partial \hat \xi_i} \lambda \equiv  M_{i,i}\zeta(f_L(x_i^L))
\end{equation}

For indices $i$ corresponding to low-fidelity data on $X_H$ ($i=n_l+1...n_l+n_h$):
\begin{equation}
\label{eq:MFGPC_partial_q_LH}
\begin{split}
\left(M_{i,i} \frac{\partial^3}{\partial \hat \xi_i ^3}  +  2M_{i,i+n_h}\frac{\partial^3}{\partial \hat \xi_{i+n_h} \partial \hat \xi_i^2} + M_{i+n_h,i+n_h}\frac{\partial^3}{\partial \hat \xi_{i+n_h} ^2\partial \hat \xi_i }\right) \lambda &\equiv \\ 
\equiv \left(M_{i,i}\rho^3 + 2M_{i, i+n_h}\rho^2 + M_{i+n_h, i+n_h}\rho\right)\zeta(\rho f^L(x^H_{i-n_l}) + \delta_{i-n_l}) &
\end{split}
\end{equation}

For indices $i$ corresponding to delta on $X_H$ ($i=n_l+n_h+1...n_l+2n_h$):
\begin{equation}
\label{eq:MFGPC_partial_q_delta}
\begin{split}
\left(M_{i,i} \frac{\partial^3}{\partial\hat  \xi_i ^3} + 2M_{i,i-n_h}\frac{\partial^3}{\partial\hat  \xi_{i-n_h} \partial\hat  \xi_i^2} + M_{i-n_h,i-n_h}\frac{\partial^3}{\partial\hat  \xi_{i-n_h} ^2\partial\hat  \xi_i }\right) \lambda &\equiv \\ 
\equiv \left(M_{i,i} + 2M_{i, i-n_h}\rho + M_{i-n_h, i-n_h}\rho^2\right)\zeta(\rho f^L(x^H_{i-n_l-n_h}) + \delta_{i-n_l-n_h}) &
\end{split}
\end{equation}

\subsection{Specifications Of Implementation}
\label{seq:specifications_of_implementation}

For experiments we used Python 3.6. 

\begin{itemize}
    \item Implementations of Gaussian Process Classifiers and Logistic Regressions were used from scikit-learn package\footnote{\texttt{http://scikit-learn.org/}}; Out method was implemented on top of \texttt{GaussianProcessClassifier} module from this package;
    \item Classifiers based on GPs used isotropic RBF kernels;
    \item Classifiers based on GPs and Logistic Regression were used in the pipeline with the Standard Scaler features preprocessor;
    \item Implementation of Gradient Boosting Classifier was used from XGBoost module\footnote{\texttt{https://xgboost.readthedocs.io}} with the following parameters: \texttt{n\_estimators=100}, \texttt{max\_depth=3}, \texttt{learning\_rate=0.05}, \texttt{subsample=0.85};
    \item For each run of the evaluation procedure we generated a random training subsample that has at least one label of each class (both for low- and high-fidelity subsamples), that is, positive and negative; 
    \item The set of random seeds for different runs was shared across series of method-dataset evaluations. 
\end{itemize}


\section{Supplementary materials}
\label{sec:supplementary}

This appendix contains supplementary materials for experimental results.

\begin{table*}[!ht]
    \centering
    \caption{Average \ROCAUC{} among multiple runs on artificial datasets from group 1.}
    \label{tab:results_group_1}
    \begin{adjustbox}{center}
    \begin{tabular}{l|cccc|cccc}
    \hline
    Noise level & \multicolumn{4}{c|}{0.2} & \multicolumn{4}{c}{0.4} \\
    \hline
    Dimensionality & 2D & 5D & 10D & 20D & 2D & 5D & 10D & 20D \\
    \hline
    \mfgpc{} & $\mathbf{0.975}$ & $\mathbf{0.853}$ & $\mathbf{0.716}$ & 0.643 & $\mathbf{0.968}$ & $\mathbf{0.750}$ & 0.615 & 0.573 \\
    \gpc{} & $\mathbf{0.970}$ & 0.732 & 0.616 & 0.587 & $\mathbf{0.970}$ & 0.732 & 0.616 & $\mathbf{0.587}$ \\
    \logit{} & 0.738 & 0.590 & 0.559 & 0.559 & 0.738 & 0.590 & 0.559 & 0.559 \\
    \xgb{} & 0.914 & 0.662 & 0.591 & 0.574 & 0.914 & 0.662 & 0.591 & 0.574 \\
    \concMF{} \gpc{} & 0.944 & $\mathbf{0.854}$ & $\mathbf{0.721}$ & $\mathbf{0.654}$ & 0.811 & 0.683 & $\mathbf{0.626}$ & $\mathbf{0.592}$ \\
    \concMF{} \logit{} & 0.721 & 0.619 & 0.580 & 0.585 & 0.675 & 0.584 & 0.557 & 0.557 \\
    \concMF{} \xgb{} & 0.916 & 0.725 & 0.644 & 0.607 & 0.807 & 0.637 & 0.586 & 0.567 \\
    \stackedMF{} \gpc{} & 0.949 & 0.812 & 0.686 & 0.616 & 0.938 & 0.713 & 0.617 & $\mathbf{0.589}$ \\
    \stackedMF{} \logit{} & 0.740 & 0.592 & 0.563 & 0.559 & 0.742 & 0.591 & 0.561 & 0.559 \\
    \stackedMF{} \xgb{} & 0.921 & 0.700 & 0.608 & 0.583 & 0.914 & 0.657 & 0.591 & 0.575 \\
    \hetmogp{} & 0.909 & 0.500 & 0.500 & 0.500 & 0.802 & 0.500 & 0.500 & 0.500 \\
    \hline
    \end{tabular}
    \end{adjustbox}
    
\end{table*}

\begin{table*}[!ht]
    \centering
    \caption{Average \ROCAUC{} among multiple runs on datasets from group 2 with noise level 0.2.}
    \label{tab:results_group_2_noise_0_2}
    
    \begin{adjustbox}{center}
    \begin{tabular}{l|ccccccccc}
    \hline
     & \diabetes{}  & \german{}  & \satimage{}  & \mushroom{}  & \splice{}  & \spambase{}  & \hypothyroid{}  & \waveform{} \\
    \hline
    \mfgpc{} & $\mathbf{0.805}$ & 0.702 & $\mathbf{0.997}$ & $\mathbf{0.997}$ & 0.936 & 0.925 & 0.646 & $\mathbf{0.919}$ \\
    \gpc{} & 0.778 & 0.704 & $\mathbf{0.997}$ & $\mathbf{0.995}$ & 0.901 & 0.907 & 0.633 & 0.908 \\
    \logit{} & $\mathbf{0.812}$ & 0.683 & $\mathbf{0.998}$ & $\mathbf{0.994}$ & 0.913 & 0.915 & 0.772 & 0.858 \\
    \xgb{} & 0.742 & 0.702 & 0.982 & 0.987 & $\mathbf{0.971}$ & 0.925 & $\mathbf{0.827}$ & 0.886 \\
    \concMF{} \gpc{} & $\mathbf{0.804}$ & 0.699 & $\mathbf{0.996}$ & $\mathbf{0.995}$ & 0.937 & 0.914 & 0.570 & $\mathbf{0.910}$ \\
    \concMF{} \logit{} & 0.803 & 0.704 & $\mathbf{0.989}$ & 0.955 & 0.794 & 0.859 & 0.654 & 0.820 \\
    \concMF{} \xgb{} & 0.767 & 0.696 & 0.987 & 0.987 & 0.958 & $\mathbf{0.946}$ & 0.791 & 0.891 \\
    \stackedMF{} \gpc{} & $\mathbf{0.804}$ & $\mathbf{0.725}$ & $\mathbf{0.997}$ & $\mathbf{0.997}$ & 0.915 & 0.914 & 0.616 & $\mathbf{0.918}$ \\
    \stackedMF{} \logit{} & $\mathbf{0.812}$ & 0.684 & $\mathbf{0.997}$ & $\mathbf{0.994}$ & 0.924 & 0.923 & 0.766 & 0.861 \\
    \stackedMF{} \xgb{} & 0.738 & 0.687 & 0.971 & 0.983 & $\mathbf{0.967}$ & $\mathbf{0.943}$ & 0.766 & 0.895 \\
    \hetmogp{} & 0.500 & 0.500 & 0.500 & 0.500 & 0.500 & 0.500 & 0.500 & 0.500 \\

    \hline
    \end{tabular}
    \end{adjustbox}
\end{table*}

\begin{table*}[!ht]
    \centering
    \caption{Average \ROCAUC{} among multiple runs on datasets from group 2 with noise level 0.4.}
    \label{tab:results_group_2_noise_0_4}
    \begin{adjustbox}{center}
    \begin{tabular}{l|ccccccccc}
    \hline
     & \diabetes{}  & \german{}  & \satimage{}  & \mushroom{}  & \splice{}  & \spambase{}  & \hypothyroid{}  & \waveform{} \\
    \hline
    \mfgpc{} & 0.781 & $\mathbf{0.710}$ & $\mathbf{0.997}$ & $\mathbf{0.996}$ & 0.905 & 0.914 & 0.676 & $\mathbf{0.909}$ \\
    \gpc{} & 0.778 & $\mathbf{0.704}$ & $\mathbf{0.997}$ & $\mathbf{0.995}$ & 0.901 & 0.907 & 0.633 & $\mathbf{0.908}$ \\
    \logit{} & $\mathbf{0.812}$ & 0.683 & $\mathbf{0.998}$ & $\mathbf{0.994}$ & 0.913 & 0.915 & 0.772 & 0.858 \\
    \xgb{} & 0.742 & 0.702 & 0.982 & $\mathbf{0.987}$ & $\mathbf{0.971}$ & $\mathbf{0.925}$ & $\mathbf{0.827}$ & 0.886 \\
    \concMF{} \gpc{} & 0.685 & 0.642 & 0.986 & 0.981 & 0.846 & 0.852 & 0.479 & 0.840 \\
    \concMF{} \logit{} & 0.743 & 0.630 & 0.934 & 0.827 & 0.626 & 0.725 & 0.579 & 0.711 \\
    \concMF{} \xgb{} & 0.697 & 0.621 & 0.934 & 0.921 & 0.831 & 0.849 & 0.674 & 0.771 \\
    \stackedMF{} \gpc{} & 0.791 & $\mathbf{0.711}$ & $\mathbf{0.997}$ & $\mathbf{0.996}$ & 0.901 & 0.906 & 0.622 & $\mathbf{0.907}$ \\
    \stackedMF{} \logit{} & $\mathbf{0.811}$ & 0.683 & $\mathbf{0.997}$ & $\mathbf{0.994}$ & 0.914 & 0.916 & 0.771 & 0.858 \\
    \stackedMF{} \xgb{} & 0.747 & 0.680 & 0.984 & $\mathbf{0.988}$ & $\mathbf{0.972}$ & $\mathbf{0.927}$ & 0.752 & 0.885 \\
    \hetmogp{} & 0.500 & 0.500 & 0.500 & 0.500 & 0.500 & 0.500 & 0.500 & 0.500 \\
    \hline

    \end{tabular}
    \end{adjustbox}
    
\end{table*}







\end{document}